\documentclass[a4paper, 10 pt, conference]{svproc}

\usepackage[caption=false,font=footnotesize]{subfig}
\usepackage[export]{adjustbox}
\usepackage{multicol}
\usepackage{footmisc}
\usepackage{float}
\usepackage{amsfonts}
\usepackage[utf8]{inputenc}

\usepackage{amsmath}
\usepackage{enumitem}
\usepackage{algorithm}
\usepackage[noend]{algpseudocode}

\begin{document}
\mainmatter

\title{Sampling-based Motion Planning via Control Barrier Functions}
\titlerunning{CBF-RRT}
\author{Guang Yang\inst{1} \and Bee Vang\inst{1} \and Zachary Serlin \inst{1} \and Calin Belta\inst{1} \and Roberto Tron\inst{1}}
\authorrunning{Guang Yang et al.}

\institute{$^1$Boston University, Boston MA, USA}
\maketitle

\begin{abstract}
Robot motion planning is central to real-world autonomous applications, such as self-driving cars, persistence surveillance, and robotic arm manipulation. One challenge in motion planning is generating control signals for nonlinear systems that result in obstacle free paths through dynamic environments. In this paper, we propose Control Barrier Function guided Rapidly-exploring Random Trees (CBF-RRT), a sampling-based motion planning algorithm for continuous-time nonlinear systems in dynamic environments. The algorithm focuses on two objectives: efficiently generating feasible controls that steer the system toward a goal region, and handling environments with dynamical obstacles in continuous time. We formulate the control synthesis problem as a Quadratic Program (QP) that enforces Control Barrier Function (CBF) constraints to achieve obstacle avoidance. Additionally, CBF-RRT does not require nearest neighbor or collision checks when sampling, which greatly reduce the run-time overhead when compared to standard RRT variants.
\end{abstract}

\section{Introduction}
Motion planning is a cornerstone of modern robotics, but it is still a challenging problem when non-trivial robot dynamics are combined with input constraints and dynamic obstacles. Control Barrier Functions (CBFs) have been shown to be effective for feedback control in similar settings; however, they cannot be used alone for producing \emph{complete} planners, as they can get trapped in ``local minima'' created by obstacles. CBFs can also become infeasible if there is no possible control that allows the robot to remain safe (i.e., to avoid collision). We look to overcome CBF limitations by generating and connecting many smaller, feasible trajectories through sampling. This results in a complete CBF-based planner that can find trajectories not considered by CBFs alone.

Sampling-based motion planning is not a novel concept, however, much of the literature is this area tends to focus on either high-level path planning, or lower level control and trajectory planning. In this paper, we attempt to present a unified approach that takes advantage of both collision free and efficient trajectory planning, as well as efficient path planning. Using CBFs developed in the formal methods and controls community, we augment the classic rapidly exploring random trees algorithm to handle dynamic obstacles and non-linear robot dynamics while also generating controls that are guaranteed to produce safe, collision free, trajectories.

\subsection{Related Work}
This work draws in past work on both sampling-based algorithms in the motion planning literature, as well as provably correct control synthesis in the formal methods for robotics community.  

Motion planning in real-world applications often considers high level path planning and low level control synthesis, given safety requirements and dynamical constraints. Sampling-based motion planning algorithms, such as Probabilistic Road Map \cite{kavraki1994probabilistic}, Rapidly-exploring Random Trees (RRT) \cite{lavalle1998rapidly}, and RRT*\cite{karaman2011sampling}, have been widely explored and are efficient strategies for high dimensional kinematic planning; however, generally these algorithms assume that a low level controller exists to generate collision free trajectories at run time. In recent years, there has been considerable effort to try to bridge the gap between path planning and control synthesis by designing controller that steer the system in-between two generated vertices, such as Kinodynamic RRT* \cite{webb2013kinodynamic}, LQR-RRT* \cite{perez2012lqr} and its variants in \cite{goretkin2013optimal} and \cite{schouwenaars2004receding}, however these approaches are limited to linear systems and static environments. Our proposed CBF-RRT method takes a different approach compared to the work in \cite{webb2013kinodynamic}, \cite{perez2012lqr}, and \cite{goretkin2013optimal}. Instead of first generating straight paths with a fixed number of vertices and then treating control synthesis as a two point boundary value problem (as in \cite{perez2012lqr}), we generate both controls and paths ``on-the-fly" and work with nonlinear systems. Motion planning in dynamical environments has been studied in \cite{adiyatov2017novel} and \cite{connell2017dynamic}, but focuses on replanning when obstacles cause collisions.

 Considerable work has been done on generating control strategies for safety-critical systems using CBFs. A popular formulation is to combine CBFs and Control Lyapunov Functions (CLFs) in a Quadratic Program (QP) where the CBF ensures safety and the CLF ensures stability. This approach has been successfully applied to applications such as adaptive cruise control \cite{ames2014control}, bipedal robot walking \cite{hsu2015control}, and swarm control \cite{borrmann2015control}. The QP based formulation works well when the desired equilibrium point is well-defined and there exists a feasible control sequence for the given problem. Under certain conditions, however, the QP may be infeasible due to the environment or dynamical constraints. Work has also been done, in \cite{herbert2017fastrack}, in designing a safe controller that follows a pre-planned path, but does not include a path-planner that explicitly explores the environment, and therefore also, occassionally, leads to infeasibility in finding a solution. 
 
\paragraph{Contributions:}
In this paper, we propose Control Barrier Function guided Rapidly-exploring Random Trees, a motion planning algorithm that uses sampling techniques to explore the state space and a QP based controller with CBF constraints to generate intermediate controls and trajectories between samples. From the sampling-based motion planning point of view, this paper provides formal guarantees for collision free continuous trajectories between samples. From a formal methods in robotics point of view, this paper offers a partial solution to infeasibility in finding a solution in CBF/CLF QP based controller formulations. The proposed framework guarantees safety, handles obstacles with known dynamics, and its internal control synthesis can be utilized as the low level controller at run-time. 

This paper is organized as follows: preliminary information is provided in Section \ref{sec:Prelim};
the problem statement and proposed algorithm are presented in Section \ref{sec:probStatement}
and Section \ref{sec:safeRRT}, respectively; a differential drive model is considered in Section \ref{sec:exampleStudy} followed by simulations in Section \ref{sec:simulation}; Conclusions are given in Section \ref{sec:conclusion}.

\section{Preliminaries}\label{sec:Prelim}
\subsection{Notation}
Let $\mathbb{R}^n$ be the set of real numbers in $n$ dimensions. The Lie derivative of a smooth function $h(x(t))$ along dynamics $\dot{x}(t)=f(x(t))$ is denoted as $\pounds_{f} h(x) := \frac{\partial h(x(t))}{\partial x(t)} f(x(t))$. Given a continuously differentiable function $h:\mathbb{R}^n \mapsto \mathbb{R}$, we denote $h^{r_b}$ as its $r_b$-th derivative with respect to time $t$. A continuous function $\alpha:(-b,a) \mapsto (-\infty,\infty)$, for some $a,b>0$, belongs to the extended class $\mathcal{K}$ if $\alpha$ is strictly increasing and $\alpha(0)=0$. The set difference between set $A$ and set $B$ is denoted as $A \backslash B$. 

\subsection{Dynamics}\label{sec:envir}
We consider a motion planing problem for a continuous-time control system
\begin{equation}\label{eq:dynamicSystem}
\dot{x} = f(x) + g(x)u,
\end{equation} where $x \in \mathcal{X} \subset \mathbb{R}^n$ is the state and $u \in \mathcal{U} \subset \mathbb{R}^m$ is the control input, where $\mathcal{U}$ is a set of admissible controls for system \eqref{eq:dynamicSystem}. The functions $f(x)$ and $g(x)$ are assumed to be locally Lipschitz continuous. The initial state is denoted as $x_{init}:=x(t_0) \in \mathcal{X}$ and the goal region is defined as $\mathcal{X}_{goal} \subset \mathcal{X}$. Obstacles are assumed to be non-stationary with known dynamics and move according to the equation
\begin{equation} \label{eq:obsDynamics}
    \dot{x}_{obs} = f_{obs}(x_{obs}),
\end{equation}
where $x_{obs} \in \mathcal{X}_{obs} \subset \mathcal{X}$ is the state variable for the obstacles (e.g., center of mass). The function $f_{obs}$ is again assumed to be locally Lipschtiz continuous. 

\begin{remark}
For special cases where the obstacle dynamics are unknown, we could approximate their dynamics with techniques, such as Gaussian Process (GP)\cite{rasmussen2003gaussian} or online Linear Regression\cite{neter1989applied}. For simplicity, in this paper, we assume that all obstacle dynamics are known. 
\end{remark}

\subsection{Exponential Control Barrier Functions (ECBF)}\label{sec:cbf}
In traditional path planning approaches, obstacle avoidance is often enforced by a collision check along segments or regions of the path. To create a more complete and provably safe trajectory, we instead formulate obstacle avoidance as remaining within a safety set defined by Control Barrier Functions (CBFs, \cite{ames2014control}) and their extensions for higher relative degree, Exponential CBFs \cite{nguyen2016exponential}. Given a continuously differentiable function $h: \mathbb{R}^n \mapsto \mathbb{R}$, the safety set $C$ is defined as
\begin{equation}\label{eq:safetySet}
\begin{aligned}
C &= \{x \in{\mathbb{R}^n}|h(x)  \geq 0 \}. \\
\partial{C} &= \{x \in \mathbb{R}^n|h(x)  = 0 \}, \\
Int(C) &= \{x \in \mathbb{R}^n|h(x)  > 0 \}
\end{aligned}
\end{equation}
where $\partial{C}$ is the boundary and $Int(C)$ is the interior.
The set $C$ is called \emph{forward invariant} for system (\ref{eq:dynamicSystem})
if $x_0 \in C$ implies $x(t)\in C$ for all $t$. 

For systems with relative degree $r_b$, we define a traverse variable as $\xi_b(x) = \left[ \begin{matrix} h(x),\dot{h}(x),\ldots,h^{(r_b)}(x) \end{matrix} \right]^T = \left[ \begin{matrix} h(x),\pounds_{f} h(x),\ldots,\pounds_{f} h^{r_b}(x) \end{matrix} \right]^T$, and formulate a virtual system with input-output linearization \cite{isidori2013nonlinear}:
\begin{align}\label{eq:ioLinearizedSys}
&\dot{\xi}_b(x) = A_b \xi_b(x) + B_b \mu, \\ \nonumber
&h_l(x) = C_b \xi_b(x),
\end{align}
where 
\begin{align*}
A_b = \left[\begin{matrix}0 &1 &\cdot &\cdot &0 \\
\cdot &\cdot &\cdot &\cdot &\cdot\\
0 &0 &0 &\cdot &1\\
0 &0 &0 &\cdot &0 \end{matrix}\right], B_b = \left[\begin{matrix}0\\
\vdots\\
0\\
1\end{matrix}\right],\\
C_b = \left[1 \ldots 0\right],
\end{align*}
and $\mu = (\pounds_{g}\pounds_{f}^{r_b-1}h(x))^{-1}(\mu-\pounds_{f}^{r_b}h(x))$ is the input-output linearized control. 

\begin{definition}[Exponential Control Barrier Function \cite{nguyen2016exponential}] Consider the dynamical system in \eqref{eq:dynamicSystem} and the safety set $C$ defined in \eqref{eq:safetySet}. A continuously differentiable function $h(x)$ with relative degree $r_b \geq 1$ is an Exponential Control Barrier Function (ECBF) if there exists $K_b \in \mathbb{R}^{1\times r_b}$, such that
\begin{equation} \label{eq:ecbfConstraint}
\inf \limits_{u\in U} [\pounds_{f}^{r_b} h(x)+\pounds_{g}\pounds_{f}^{r_b-1} h(x)u+ K_b \xi_b(x) ] \geq 0, \forall x \in Int(C),
\end{equation}
where the row vector $K_b$ is selected such that the closed-loop matrix $A_b -B_b K_b$ for \eqref{eq:ioLinearizedSys} is stable.
\end{definition}

\begin{theorem}
Given the system \eqref{eq:dynamicSystem}, and the safety set $C$ defined in \eqref{eq:safetySet}, if there exists an ECBF $h(x)$, then the  system is forward invariant in $C$ \cite{nguyen2016exponential}.
\end{theorem}

\begin{remark}\label{ECBFremark}
There exist a more general notion of  Higher Order Control Barrier Function (HOCBF) \cite{xiao2019control}, of which ECBF is a special case. For the purposes of this paper, we only need ECBFs, although the use of HOCBFs could lead to improved performance.
\end{remark}

\subsection{Rapidly-exploring Random Trees (RRT)}\label{sec:classicRRT}
 We use $\mathcal{T} = (\mathcal{V}, \mathcal{E})$ to denote a tree with a set of vertices $\mathcal{V} \subseteq \mathcal{X}$ and a set of edges $\mathcal{E}$. The classic Rapidly-exploring Random Trees algorithm (RRT, \cite{lavalle1998rapidly}), builds a tree $\mathcal{T}$ where each vertex $\mathit{v} \in \mathcal{V}$ is associated to a full state $\mathit{v} = x \in \mathbb{R}^n$, and edges follow the system dynamics \eqref{eq:dynamicSystem}. To work, the algorithm needs the following essential components:
\begin{itemize}
 \item \textbf{State Space and Goal Region}: A configuration space $\mathcal{X}$, such that $x\in \mathcal{X}$, and a goal region $\mathcal{X}_{goal} \subset \mathcal{X}$.
 \item \textbf{Collision Check}: A function $\texttt{CollisionCheck}(x)$ that detects if a state trajectory $x$ violates any collision constraint. 
 \item \textbf{Metric}: A function $d:\mathcal{X}\times\mathcal{X} \rightarrow [0,\infty)$ that returns the distance between two vertices within $\mathcal{X}$.
 \item \textbf{Nearest Neighbor}: A function $\texttt{NearestNeighbor}:(\mathcal{T}, x_{sample})\mapsto x_{nn}$ which utilizes the distance function $d$ to find the vertex $x_{nn}$ in $\mathcal{T}$ that is closest to $x_{sample}$.
 \item \textbf{Inputs}: A set of admissible controls $U$, such that $u\in U$, for steering the state $x$.
 \item \textbf{Steer}: Given a sampled vertex $x_{sample}$ and nearest vertex $x_{nn}$, the function $\texttt{Steer}(x_{nn},x_{sample})$ steers the system from $x_{nn}$ to $x_{sample}$ with control $u$ while performing collision checks. The function terminates after the trajectory progresses some fixed distance and returns false if no trajectory is possible. 
\end{itemize}
The RRT algorithm is summarized in Algorithm \ref{algorithm0} and Algorithm \ref{algorithm1}.

\begin{algorithm} 
\caption{RRT}\label{algorithm0}
\begin{algorithmic}
    \State $\mathcal{T} = (\mathcal{V},\mathcal{E})$, $\mathcal{V} \leftarrow$ $\{x_{init}\}$; $\mathcal{E} \leftarrow \emptyset$ ;
    \While{$x \not\in \mathcal{X}_{goal}$}
        \State $x_{sample}$ $\leftarrow$ \texttt{RandomState}($\mathcal{V}$)
        \State $\mathcal{T} \leftarrow$ \texttt{Extend}($\mathcal{T}$,$x_{sample}$)
    \EndWhile
\Return $\mathcal{T}$
\end{algorithmic}
\end{algorithm}
\vspace{-15pt}
\begin{algorithm} 
\caption{Extend($\mathcal{T}$,$x_{sample}$)}\label{algorithm1}
\begin{algorithmic}
    \State $x_{nn}$ $\leftarrow$ \texttt{NearestNeighbor}($\mathcal{T},x_{sample}$);
    \If{Steer($x$,$x_{nn},x_{new},u_{new}$)} 
        \State $\mathcal{V}.\texttt{AddVertex}(x_{new})$
        \State $\mathcal{E}.\texttt{AddEdge}(x_{new}, u_{new}, x_{nn})$
        \If{$x_{new}$ = $x_{sample}$}
            \State \Return $(\mathcal{V},\mathcal{E})$
        \Else 
            \State \Return  $\mathcal{T}$
        \EndIf
    \EndIf
    \Return Trapped
\end{algorithmic}
\end{algorithm}

\begin{figure}[]
\centering
 \includegraphics[width=0.40\textwidth]{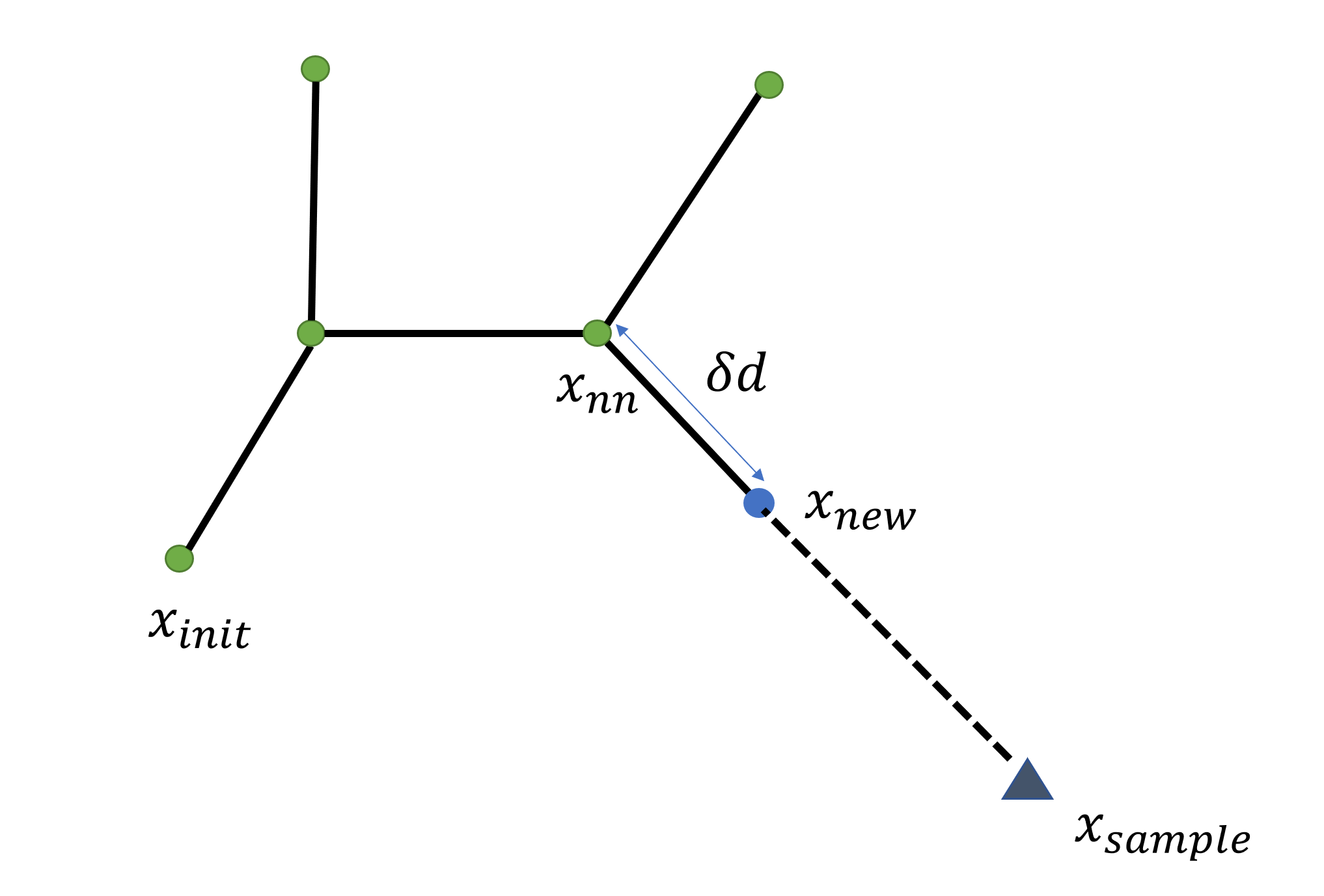}
  \caption{\texttt{Steer} function in classic RRT}\label{fig:classicRRT}
\end{figure}

 There are three possible outcomes from the \texttt{Steer} function: \texttt{Reached} indicates that $x_{new} = x_{sample}$ can be reached under the dynamical and obstacle constraints, \texttt{Advanced} indicates that $\|x_{sample} - x_{new}\| \leq \delta d$, where $\delta d$ is a positive constant, \texttt{Trapped} indicates that the function cannot find a feasible control $u$ to steer the system. The \texttt{Steer} function is demonstrated in Figure \ref{fig:classicRRT}. 
 
\section{Problem Statement and Approach}\label{sec:probStatement}
\begin{problem}
Consider a nonlinear system in the form of \eqref{eq:dynamicSystem}, with initial state $x_{init} \in \mathcal{X}_{init}  \subset \mathcal{X} \subset \mathbb{R}^n$, where $\mathcal{X}_{init}$ is an initial obstacle free set, and a bounded goal region $\mathcal{X}_{goal}$, generate feasible control inputs $u(t)$ that steer the system to $\mathcal{X}_{goal}$ while avoiding dynamical obstacles.
\end{problem}

To approach the problem, we combine both obstacle dynamics and system dynamics into a composite system. By treating the obstacles' state as part of the composite system, we can effectively construct the CBF constraints for the QP controller. The path planning algorithm is given in Section \ref{algorithm2}.

\section{CBF-RRT}\label{sec:safeRRT}
The CBF-RRT algorithm is different from other RRT variants in that there is no explicit collision or nearest neighbor checks. Instead, we introduce the notion of safe steering that encodes collision avoidance as staying within the safety set \eqref{eq:safetySet}. The nearest neighbor check is implicitly handled by sampling the vertices in $\mathcal{V}$.

Before we formally introduce the algorithm, we define the following components:
\begin{itemize}
 \item \textbf{State Space}: A topological space $\mathcal{X} \subset \mathbb{R}^n$, and $\mathcal{X}_{goal}, \mathcal{X}_{obs} \subset \mathcal{X}$.
 \item \textbf{Inputs}: A set of admissible controls $U$ for steering the state $x$.
 \item \textbf{Safe Steering}: A function that generates both safe controls and trajectories in a time interval $t_h$, given the system kinematics \eqref{eq:dynamicSystem} and obstacle dynamics \eqref{eq:obsDynamics}. More details are introduced in Section \ref{SafeSteering}.
\end{itemize}

In general, the vertices are sampled points of the state space, $\mathcal{V} \subseteq \mathcal{X}$ and we assume $\mathcal{X}_{goal} \backslash \mathcal{X}_{obs}$. To account for the time-varying obstacles, the vertices need to store an additional parameter that determines when, in time, the vertices are safe. In other words $\mathcal{V} \subseteq \mathcal{X}\times\mathbb{R}$ where the additional parameter is time and an element $v=(x,t) \in \mathcal{V}$. We use $\mathcal{T} = (\mathcal{V \subseteq \mathcal{X} \times \mathbb{R}}, \mathcal{E})$.

\subsection{Sampling}
\subsubsection{Vertices} \label{samplingInTree}
The function $\mathtt{VerticesSample}: v_s \in \mathcal{V}$ samples, with a desired probability distribution, $p_v(\mathcal{V})$, on the existing vertices of the tree $\mathcal{T}$. By varying the probability distributions, the behavior of the tree expansion, as well as convergence speed, will be drastically different.


\subsubsection{State Space}\label{samplingInSS}
Depending on the problem, some state variables may not play an essential role in the task requirements and/or system dynamical constraints. We denote these variables as free state variables that can be arbitrarily chosen to increase the probability of finding a path that satisfies the dynamics and safety constraints. We define function $\mathtt{StateSample}: v_s \mapsto v_{e}$ that updates the free state variables of a given vertex with a probability distribution. We denote this probability distribution as $p_{state}(x)$ and define $v_e$ as the expanding vertex. For example, if the task is to steer a nonholonomic first-order planar robot, such as a unicycle, from one position to another and orientation does not matter. Then, the vertices only need to contain information relevant to the position of the robot, i.e. $\mathcal{V} \in \mathbb{R}^{2}\times\mathbb{R} \subset \mathbb{R}^{3}\times\mathbb{R}$.

\subsubsection{Control Reference}
If no free states exist in the system dynamics, the $u_{ref}$ value must be sampled from a distribution $p_{u_{ref}}$ in order to maintain the probabilistic nature of CBF-RRT. In this case, we define a function $\mathtt{ReferenceSample}: u_{ref} \in U$ that updates the $u_{ref}$ for the chosen sample from $p_{u_{ref}}$.

\subsection{Safe Steering}\label{SafeSteering}
We define a function \texttt{SafeSteer}$(v_{0}, t_h, u_{ref})$ that contains two components: controls synthesis and collision-free trajectory generation. Given an initial vertex $v_0$ which includes an initial state $x_0$ and time element $t_0$, a fixed time horizon $t_h$, a control reference $u_{ref}$, obstacle dynamics \eqref{eq:obsDynamics} and system dynamics \eqref{eq:dynamicSystem}, \texttt{SafeSteer} solves a sequence of QPs (Section \ref{QPformulation}) with CBF constraints and generate a sequence of control inputs $u(t)$. The control inputs generate a collision-free trajectory to $x_{new}$ at time $t_0+t_h$ which is added to the tree $\mathcal{T}$ as a new vertex.

\begin{remark}
The control reference $u_{ref}$ determines how the robot explores the space and the QP ensures the robot's safety while doing so by modifying $u_{ref}$ as necessary. For example, if the reference command $u_{ref}$ is defined as "go forward in the body-fixed frame", then the QP's job is to steer the robot away from obstacles when necessary. If there are no obstacles, the robot should move in a straight line.
\end{remark}

\subsection{Goal Check}\label{goal}
Given a desired goal $x_{goal}$,  we define a goal region $\mathcal{X}_{goal} = \{y \in \mathbb{R}^n : d(y,x_{goal}) \leq \epsilon \}$, where $\epsilon$ is a positive constant chosen such that $\mathcal{X}_{goal}$ is obstacle free. If the trajectory $x(t) \in \mathcal{X}_{goal}$, then the algorithm terminates and a path is found. Otherwise, the algorithm continues.

\subsection{Quadratic Program Formulation with Control Barrier Function}\label{QPformulation}
The QP based controller takes in a reference control $u_{ref}$, current state of the system $x(t)$ and obstacle $x_{obs}(t)$ as inputs and finds a feasible control $u(t)$ point-wise in time that tries to follow $u_{ref}$. Given a safety set $h_i(x)$ and a system with relative degree $r_b$, we define the $i$-th CBF constraint as

\begin{equation*}
    \zeta_i(x) = \pounds_{f}^{r_b} h_i(x)+\pounds_{g}\pounds_{f}^{r_b-1} h_i(x)u+ K_{b,i} \xi_{b,i}(x) \geq 0. 
\end{equation*}
Additional linear constraints (with respect to $u$) may be considered, such as control bounds. These requirements may be formulated as a QP as follows:
\begin{equation}\label{eq:ECBFQP}
\begin{aligned}
& \underset{u\in \mathrm{U}}{\text{min}}
& & \|u-u_{ref}\|^{2}\\
& \text{s.t.}
& & \zeta_i (x) \geq 0, \quad i = 1,...,N_{obs}\\
& & &\underline{u} \leq u \leq \overline{u},\\
\end{aligned}
\end{equation}
where $\underline{u},\overline{u}$ are the lower and upper control bounds, respectively. The CBF constraint $\zeta_i$ is linear in terms of decision variable $u$ and $N_{obs}$ is the total number of obstacles.

\begin{algorithm} \label{algorithm2}
\caption{CBF-RRT}
\begin{algorithmic}[1]
    \State $\mathcal{V} \leftarrow$ $\{(x_{init},t_{init})\}$; $\mathcal{E} \leftarrow \emptyset$ ; \Comment{Initialize first vertex to initial state.} 
    \State $\mathcal{X}_{goal} \subset \mathcal{X}$ \Comment{Define goal region.}
    \While{$x \not\in \mathcal{X}_{goal}$}
        \State $v_s$ $\leftarrow$ \texttt{VerticesSample}($\mathcal{V}$, $p_v$) \Comment{Sample a vertex in $\mathcal{V}$}
        \State $v_{e}$ $\leftarrow$ \texttt{StateSample}($v_s$, $p_{state}$) \Comment{Sample the state at vertex $x_s$ (i.e. $\theta$)}
        \State $u_{ref}$ $\leftarrow$ \texttt{ReferenceSample}($p_{u_{ref}}$) \Comment{Sample reference control if needed}
        \State $\textbf{u}_{traj}$, $\textbf{x}_{traj}$, $x_{new}$, $t_{new}$ $\leftarrow$ \texttt{SafeSteer}($v_{e}$, $t_h$,$u_{ref}$) \Comment{Solve CBF QP}
        \If{$x_{new} \neq \emptyset$} \Comment{If QP was feasible}
            \State $\mathcal{V}$ $\leftarrow$ $(x_{new},t_{new})$,$\mathcal{E}$ $\leftarrow$ $\textbf{x}_{traj}$ \Comment{Update the tree}
        \EndIf
    \EndWhile  \label{rrt loop}
    \Return $\mathcal{T} = (\mathcal{V},\mathcal{E}), \textbf{u}_{traj}$
\end{algorithmic}
\end{algorithm}


\begin{algorithm} \label{algorithm3}
\caption{SafeSteer}
\begin{algorithmic}[2]
    \State Given $v_e$, $t_h$, $u_{ref}$
    \State $\zeta_i$ $\leftarrow$ i-th CBF constraint, $\forall i$;
    \State $\textbf{u}_{traj}$, $\textbf{x}_{traj}$, $x_{new}$, $t_{new}$ $\leftarrow$ \texttt{Integrator}($x_e,t_e,t_h, \texttt{QPcontroller}(x(t),x_{obs}(t),u_{ref}))$;
    \State \Return $\textbf{u}_{traj}$, $\textbf{x}_{traj}$, $x_{new}$, $t_{new}$
\end{algorithmic}
\end{algorithm}

\begin{figure}[H]
\begin{center}
\begin{tabular}{@{}c@{}}
\includegraphics[width=.4\textwidth]{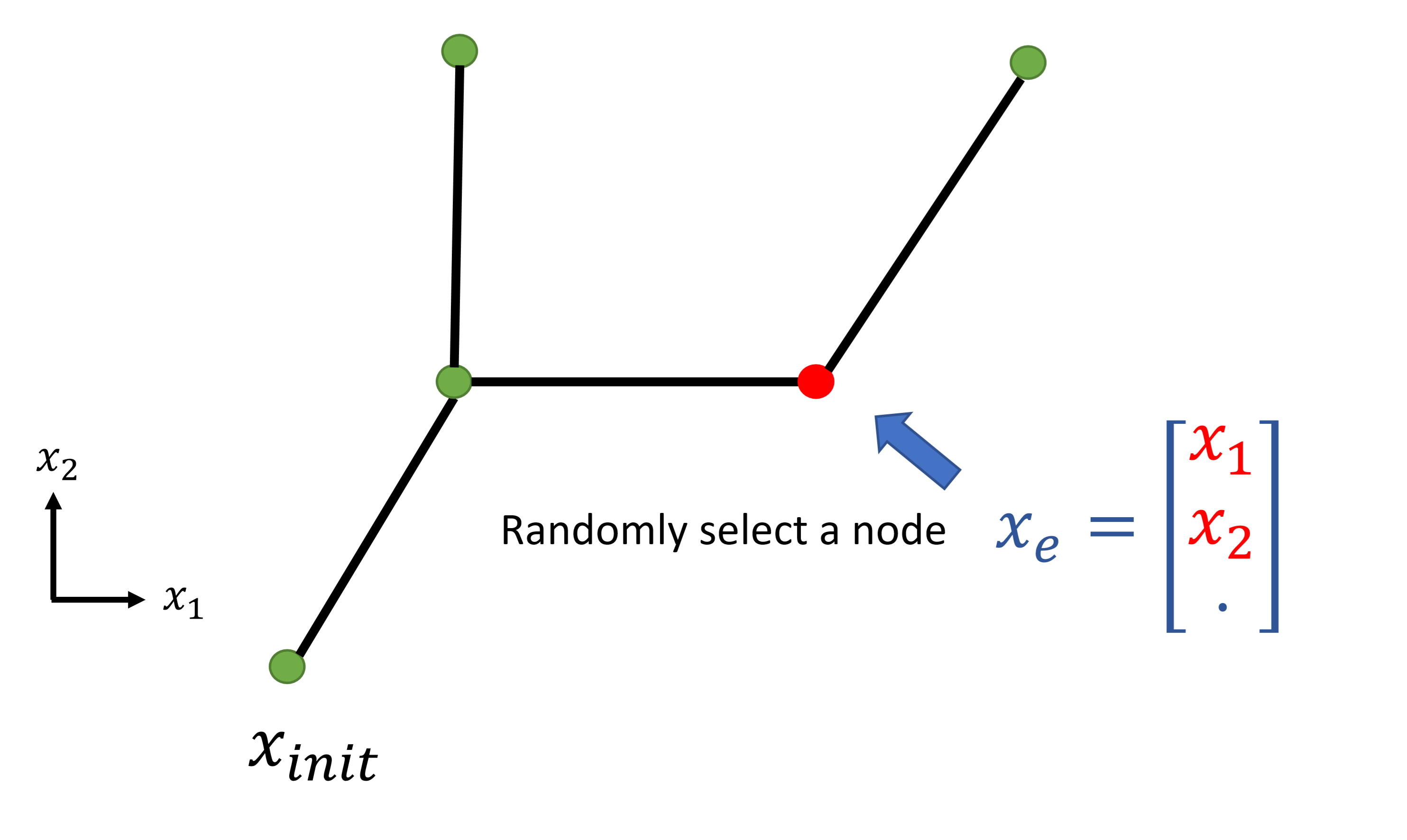}\\
\small{(a)}
\end{tabular}
\begin{tabular}{@{}c@{}}
\includegraphics[width=.4\textwidth]{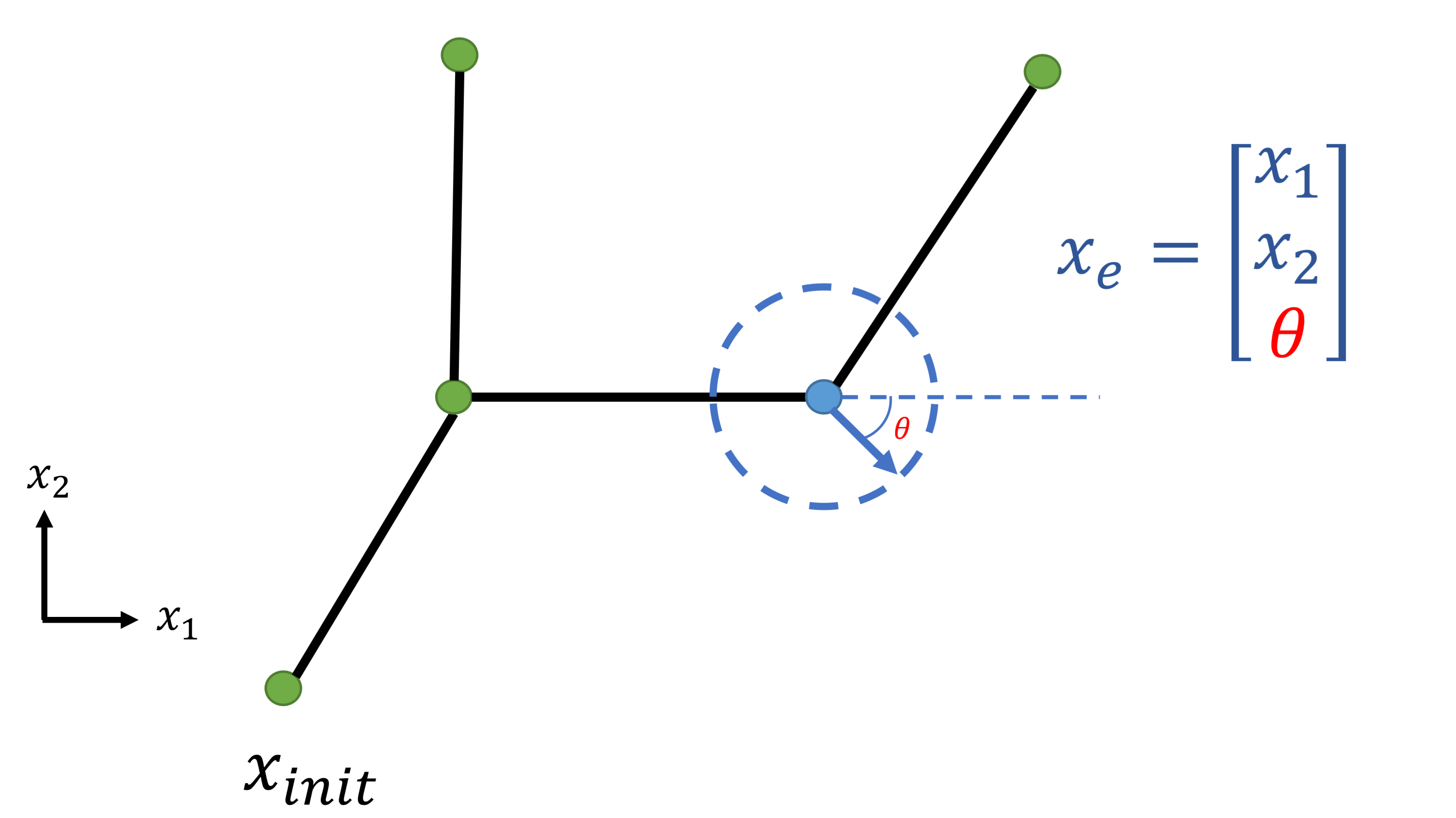}\\
\small{(b)}
\end{tabular}
\begin{tabular}{@{}c@{}}
\includegraphics[width=.4\textwidth]{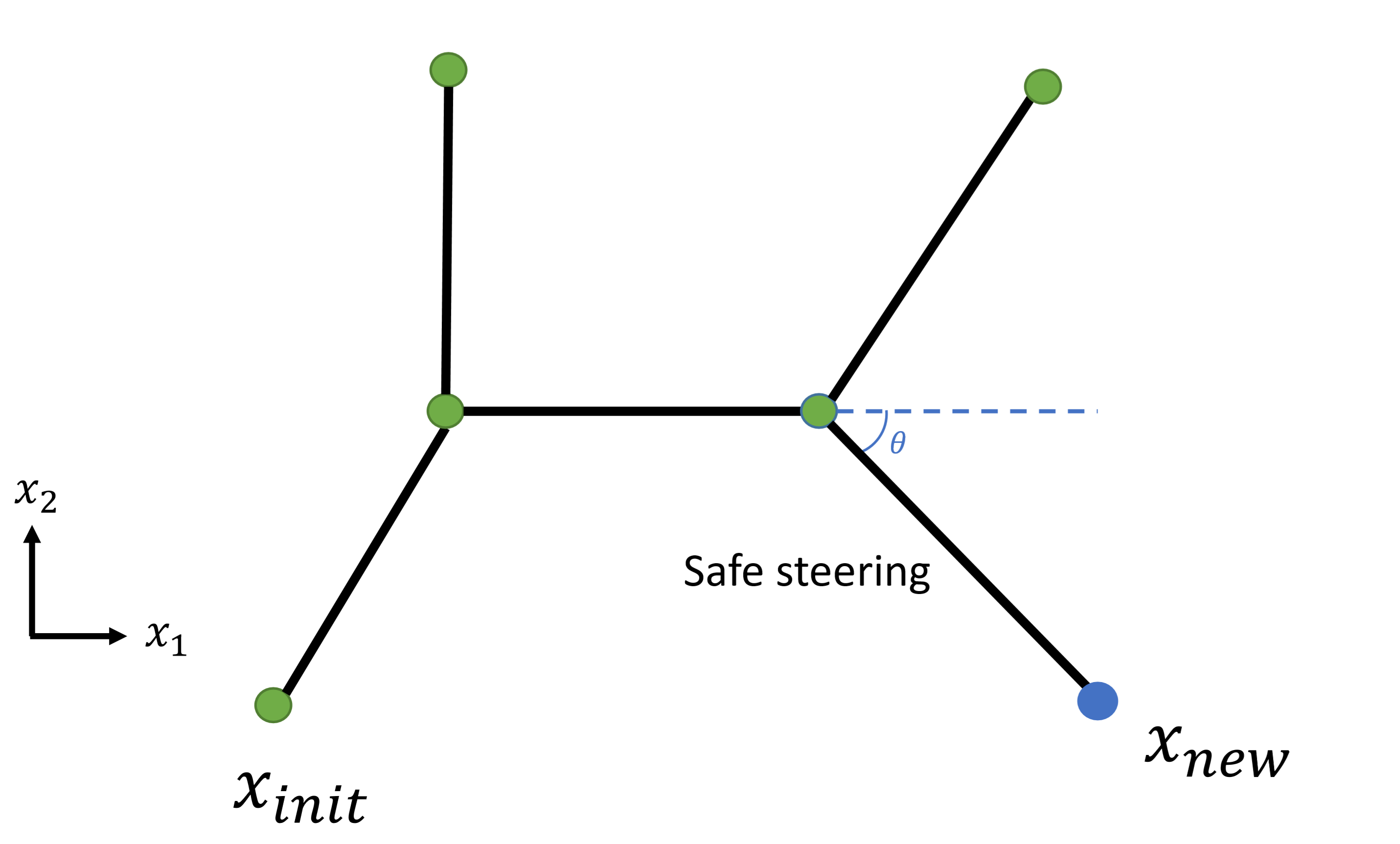}\\
\small{(c)}
\end{tabular}
\begin{tabular}{@{}c@{}}
\includegraphics[width=.4\textwidth]{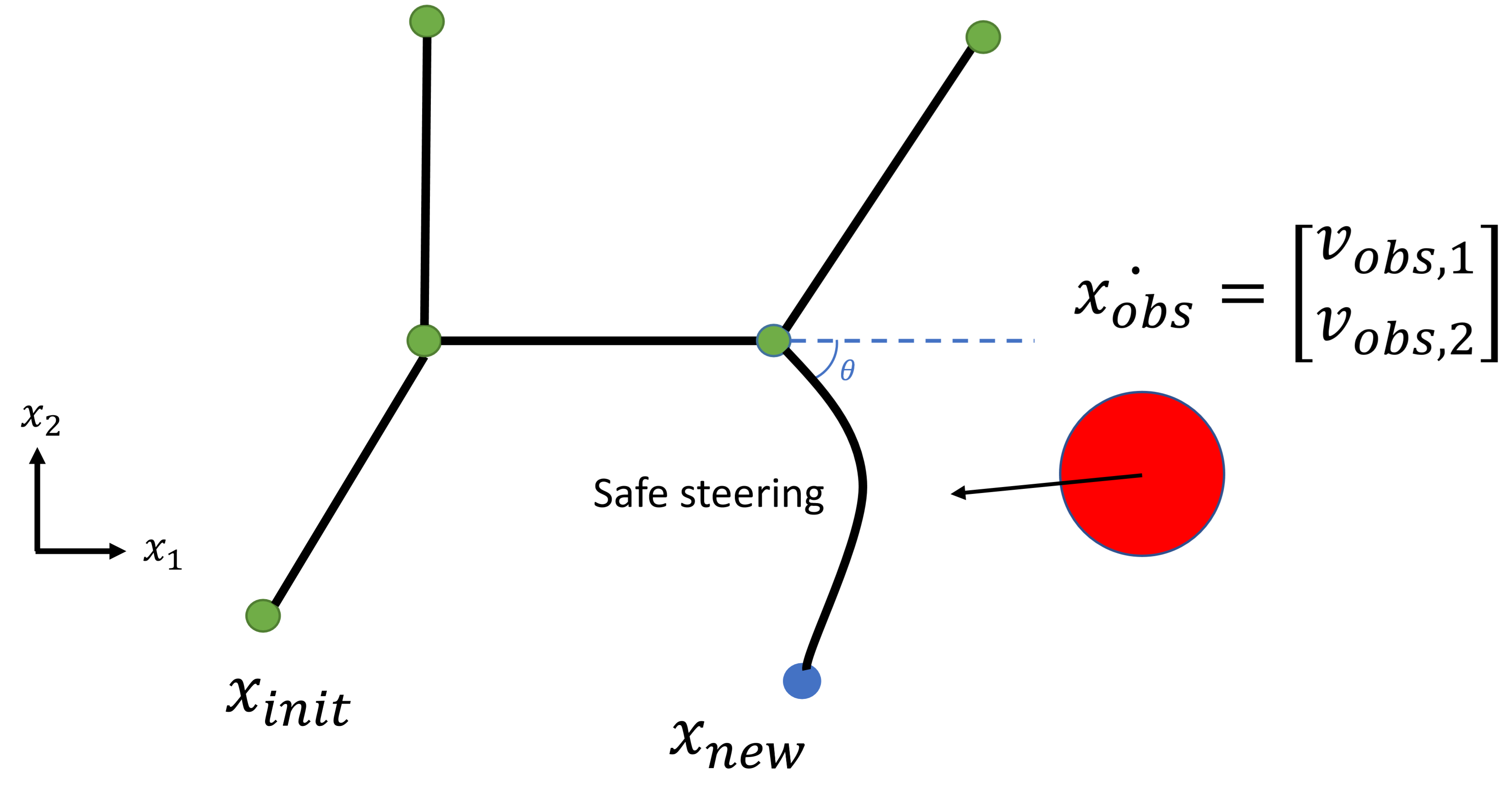}\\
\small{(d)}
\end{tabular}
\end{center}
\caption{Illustration of CBF-RRT for a system with states $x = [x_1, x_2, \theta]^T$ where $\theta$ is a free variable: The algorithm picks a random vertex within tree $\mathcal{T}$ and sets it as the sampled vertex $v_s$. Next, it performs another sampling on the state variable $\theta$ under state distribution $p_{state}$. The steering function then generates a sequence of controls $u$ to steer the system while avoiding collisions. (a) Selection of a random vertex in $\mathcal{V}$. (b) Selection of random state $\theta$. (c) Safe steering trajectory when no nearby obstacle present. (d) Safe steering trajectory with dynamic obstacle present.}
\end{figure}

\section{Numerical Examples}\label{sec:exampleStudy}
In this section, we consider the motion planning problem of steering a planar robot from an initial state $x_{init}$ to a goal position that is independent of orientation. In particular, the goal region $\mathcal{X}_{goal}$ is defined by $x_{goal}$ and $\epsilon$ 
where $x_{goal} = [x_{1,goal},x_{2,goal},\theta_{goal}]^{T}$ and $\theta_{goal}$ is arbitrary. Thus, the motion planning problem is solved in the work space $\mathcal{X}_{work} \in \mathbb{R}^{2}$.
\subsection{Dynamical System}
Consider a unicycle model for a two-wheeled differential drive robot
\begin{align} \label{eq:sysModel}
    &\dot{x_1} = v \cos(\theta), \\ \nonumber
    &\dot{x_2} = v \sin(\theta), \\ \nonumber
    &\dot{\theta} = \omega,
\end{align}
where the state $x = [x_1,x_2,\theta]^T \in \mathbb{R}^3$ corresponds to the location ($x_1,x_2$) in work space $\mathcal{X}_{work} \subset \mathbb{R}^2$ and heading $\theta$ with respect to the inertial frame. The control input $u = [v, \omega]^T \in \mathbb{R}^2$ consists of the translational and angular velocity that are bounded, respectively. The equations of motion \eqref{eq:sysModel} can be write in control affine form as
\begin{align}
    \left[\begin{matrix}\dot{x_1} \\ \dot{x_2} \\ \dot{\theta}\end{matrix} \right]= \left[ \begin{matrix} \cos(\theta)\\ \sin(\theta) \\0 \end{matrix} \right] v + \left[ \begin{matrix} 0\\ 0 \\1 \end{matrix} \right] \omega.
\end{align}

\subsection{Safety Sets}
We consider rigid body obstacles and model them as the union of circles with centroids $(x_{obs,i,1}(t),x_{obs,i,2}(t))$ and fixed radii $r_{obs,i}$ where each obstacle is inscribed by the union of their respective circles. We denote the i-th safety set as
\begin{equation}\label{eq:C_example}
    C_i = \{x \in R^2 : h_{i}(x)  \geq 0\},
\end{equation}
where 
\begin{equation}\label{eq:h_func}
    h_{i}(x) = (x_1(t) - x_{obs,i,1}(t))^2 + (x_2(t) - x_{obs,i,2}(t))^2-r_{obs,i}^2. 
\end{equation}
Each circle has the following dynamics
\begin{align} \label{eq:obsDynamicsExamples}
     &\dot{x}_{obs,i,1} = v_{obs,i,1}, \\ \nonumber
     &\dot{x}_{obs,i,2} = v_{obs,i,2}, \forall i. \\ \nonumber
\end{align}
The safe set of the robot is given as the intersection of all the safe sets for the circles
\begin{equation}
    C_{robot}=\bigcap_{i=1}^{N_{obs}} C_i.
\end{equation}

The relative degree is 1 for translation velocity $v$ and 2 for angular velocity $\omega$, with respect to system dynamics \eqref{eq:sysModel} and CBF constraint \eqref{eq:h_func}. Therefore, we have a nonlinear system with mixed-relative degree control inputs and $r_{b}=2$. The i-th ECBF constraint is
\begin{equation} \label{eq:ecbfConstraintExample1}
 \zeta_i(x) = \pounds_{f}^{2} h_i(x)+\pounds_{g}\pounds_{f} h_i(x)u+ k_1 h_i(x)+k_2 \pounds_{f}h_i(x)  \geq 0, \forall x \in Int(C_i),
\end{equation}
where 
\begin{align}
    h_{i}(x) = (x_1(t) - x_{obs,i,1}(t))^2 + (x_2(t) - x_{obs,i,2}(t))^2-r^2 \\ \nonumber
    \pounds_{f} h_{i}(x) = 2v(x_1(t) - x_{obs,i,1}(t))\cos{\theta} + 2v(x_2(t) - x_{obs,i,2}(t))\sin{\theta} \\ \nonumber - 2 (x_1(t)-x_{obs,i,1}(t))v_{obs,i,1} - 2 (x_2(t)-x_{obs,i,2}(t))v_{obs,i,2},
\end{align}
and $k_1, k_2$ are positive constants that are selected appropriately to ensure forward invariance, as mentioned in \cite{nguyen2016exponential}. The resultant ECBF constraint is
\begin{align} \label{eq:generalECBFexample}
    \zeta_i(x) = 2x_1v^2\cos^2{\theta}+2x_2v^2\sin^2{\theta}\\ \nonumber+(2(x_2-x_{obs,2})\cos{\theta}-2(x_1-x_{obs,1})\sin{\theta})\omega \\ \nonumber + 2v_{obs,i,1}^2+2v_{obs,i,2}^2+k_1 h(x) + k_2 \pounds_{f} h(x) \geq 0. 
\end{align}
\begin{remark}\label{remark:mixRelativeDegree}
The inequality \eqref{eq:generalECBFexample} is linear with respect to  $\omega$ but not $v$, therefore it cannot be add directly into the QP as a linear constraint. This is due to the mixed-relative degree of the control inputs. To overcome this limitation, we set the translational velocity $v = c \in \mathbb{R}$, which becomes part of the system dynamics function $f(x)$ in \eqref{eq:dynamicSystem}.
\end{remark}


\subsection{Sampling Distributions}
\paragraph{Vertex Sampling:} Given a set of vertices $\mathcal{V}$ in $\mathcal{T}$, we define a discrete uniform distribution over all vertices
\begin{equation}
    p_{v}= \frac{1}{N_{v}},
\end{equation}
where $N_{v}$ is the total number of vertices in $\mathcal{V}$. Therefore, each vertex has equal probability to be selected as an expanding vertex. 

\paragraph{State Sampling:} Given the state of the current sampled vertex $[x_{1,s},x_{2,s},\theta_{free}]^T$, we first calculate the desired heading angle $\theta_{goal}$ toward $x_{goal}$ by
\begin{equation*}
    \theta_{goal} = \arctan\left(\frac{x_{2,goal}-x_{2,e}}{x_{1,goal}-x_{1,e}}\right)
\end{equation*}
We then define a Gaussian distribution for the state $\theta_{free}$
\begin{equation}\label{eq:pState}
    p_{state}(\theta_{free}) = \frac{1}{\sqrt{2 \pi \sigma^2}} e^{-\frac{(\theta_{free} - \theta_{goal})^2}{2\sigma^2}},
\end{equation}
where $\sigma^2$ is the variance and we set $\theta_{goal}$ as the mean.

\section{Simulations}\label{sec:simulation}
For simulations, the control inputs are $v \in [-1,1]$ and $\omega \in [-4.25,4.25]$. We then add \eqref{eq:generalECBFexample} as linear constraints in the QP \eqref{eq:ECBFQP} with $\omega$ as the only decision variable and $v=1$ (see Remark \ref{remark:mixRelativeDegree}).
We choose $\omega_{ref} = 0$, i.e., the system minimally change its heading direction $\theta$. The combination of $(v,\omega_{ref})$ is chosen such that the robot moves in straight lines as it explores the space.  All experiments are performed on a desktop computer with a i7-8700K CPU with the Gurobi 8.1.1 \cite{gurobi} solver.  

\subsection{Example 1: Static Environment}
In the first example, we consider an environment with $N_{obs}=3$ static obstacles, i.e., $ v_{obs,i,1} = v_{obs,i,2} = 0, \forall i = 1,...,N_{obs}$, which implies $x_{obs,i,1}$ and  $x_{obs,i,2}$ are constants in \eqref{eq:h_func}. The corresponding ECBF is
\begin{align*}
    \zeta_{i}(x) &= 2 x_1 v^2 \cos^2(\theta) + 2x_2 v^2 \sin^2 (\theta) + k_1 h_{i}(x) + k_2 (2 x_1 v \cos(\theta) + 2x_2 v \sin (\theta))  \\&+[-2v(x_1-x_{obs,i,1})\sin (\theta)+ 2v(x_2-x_{obs,i,2})\cos (\theta)]w \geq 0. 
\end{align*}
The initial state is $x_{init} = [-0.5,-0.5,1]^T$ and the goal state is $x_{goal} = [2,2,\cdot]^T$. We define three obstacles with their centroids locate at $(0.3,1.2)$, $(1.0,0.5)$, $(1.7,-0.5)$ with a radius $r_i = 0.2, \forall i$. The hyper parameters in Table \ref{tab:hyperParam1} include ECBF coefficients $k_1, k_2$, variance $\sigma^2$ for state distribution $p_{state}$, radius of the goal region $\epsilon$ and time horizon $t_h$. 

\begin{figure}[H]
    \centering
    \includegraphics[width=.40\textwidth,trim={0 0 0 5.8cm},clip=True]{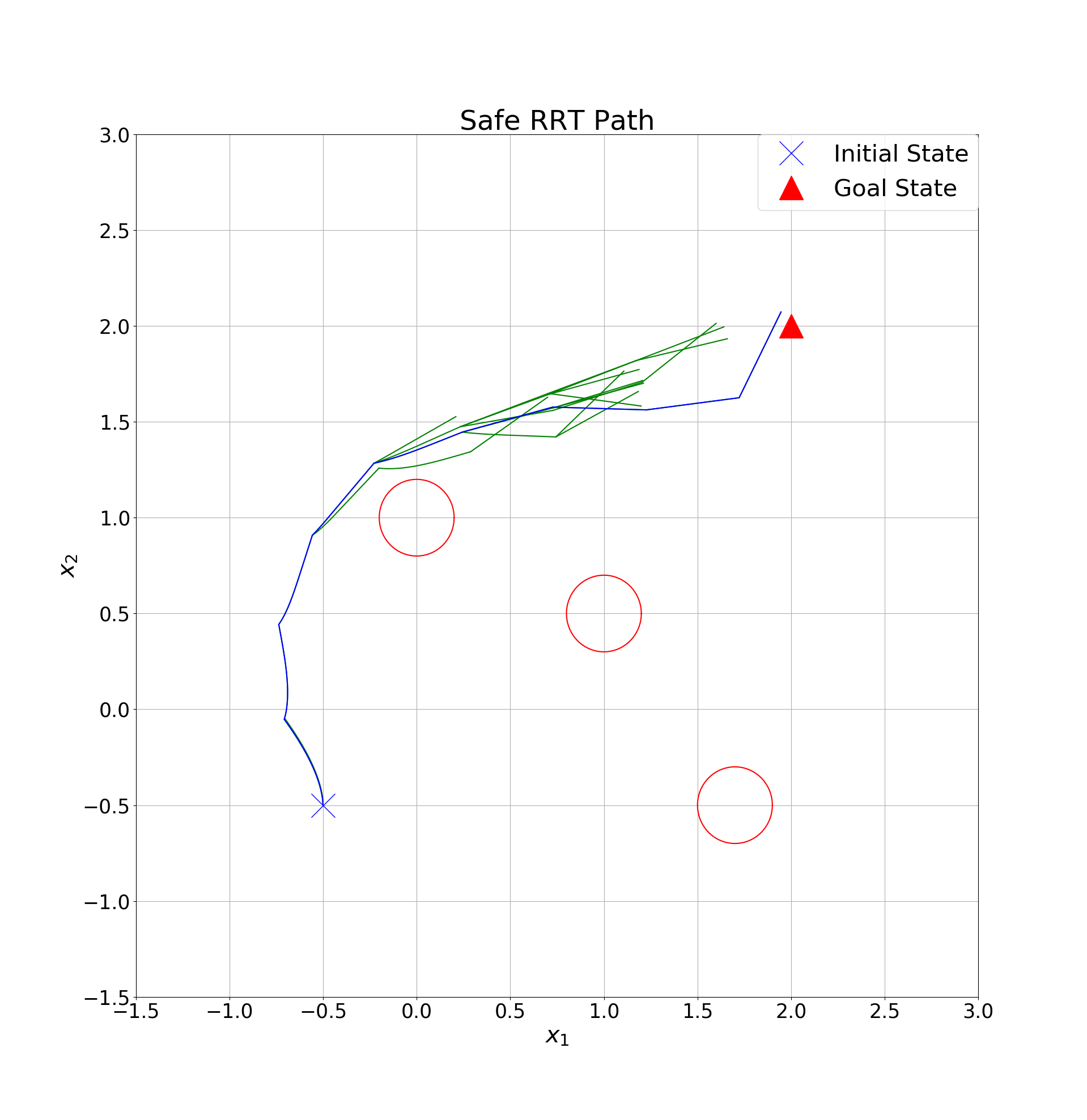}
    \includegraphics[width=.40\textwidth,trim={0 0 0 5.8cm},clip=True]{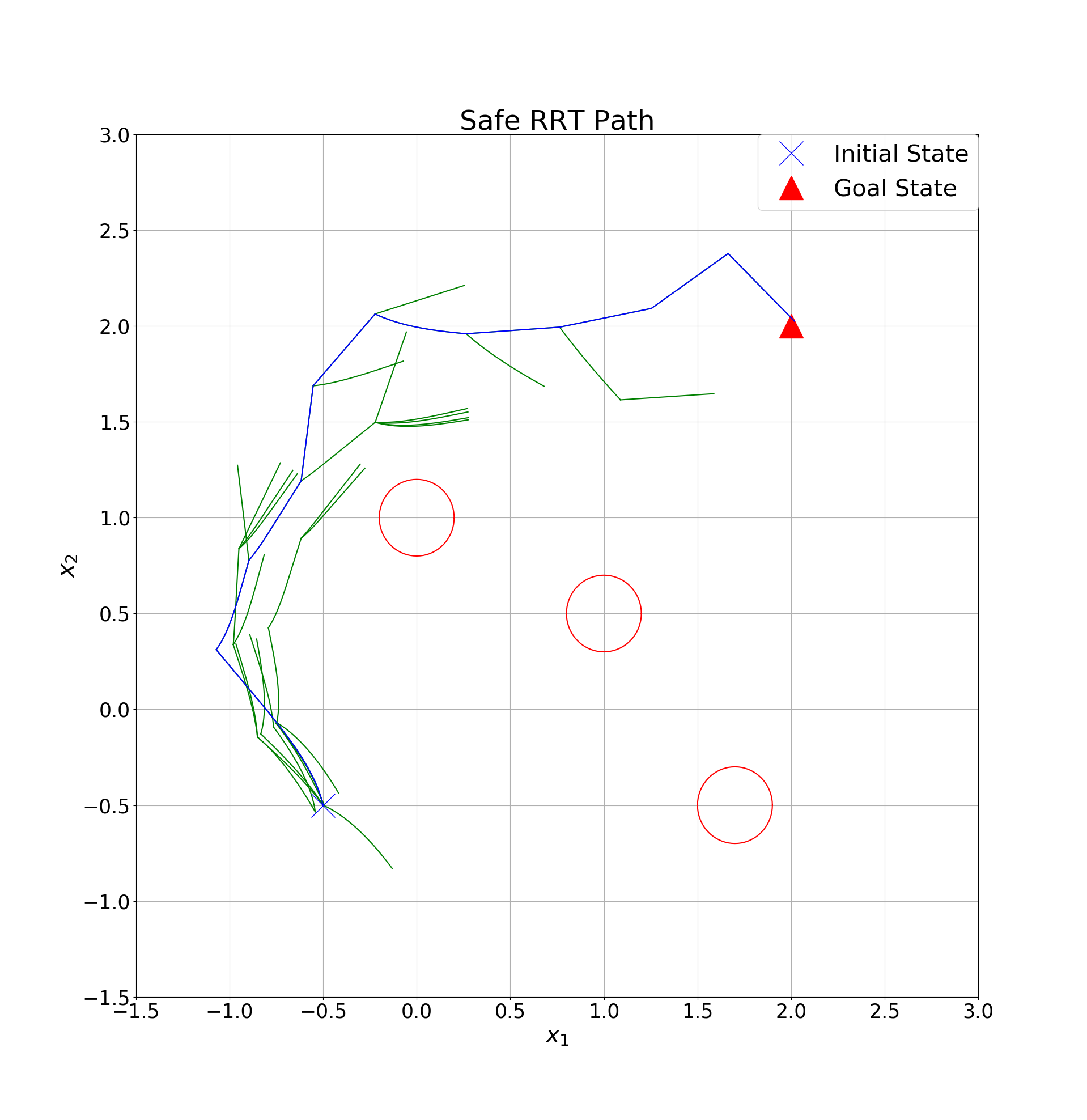}
    \caption{Example 1: CBF-RRT Motion Planning}
    \label{fig:Example1_2}
\vspace{-10pt}    
\end{figure}
\begin{table}[H]\label{tab:hyperParam1}
\centering
\caption{Hyper Parameters for Example 1}\label{tb:example1}
\begin{tabular}{ |c|c|c|c|c|c|c|c| } 
\hline
Case & $k_1$ & $k_2$ & $\sigma^2$  & $\epsilon$ & $t_h$& v & Run Time \\
\hline
1& 2.0 & 4.0 & 0.2& 0.15 &0.5 &1.0 &20.37s \\ 
 \hline
2& 2.0 & 4.0 & 0.6& 0.15& 0.5&1.0 & 1.81s \\ 
\hline
\end{tabular}
\end{table}
In Fig. \ref{fig:Example1_2}, the generated tree curves around the obstacle as the $\texttt{SafeSteer}$ function ensures the generated trajectory never enter the obstacle regions. Note case 2 has a higher variance ($\sigma^2=0.6$) which result in more exploration in the work space before finding a path to $\mathcal{X}_{goal}$. 
\subsection{Example 2: Dynamical Environment}
\begin{figure}
\begin{center}
\begin{tabular}{@{}c@{}}
\includegraphics[width=.40\textwidth,trim={0 0 0 2.7cm},clip=True]{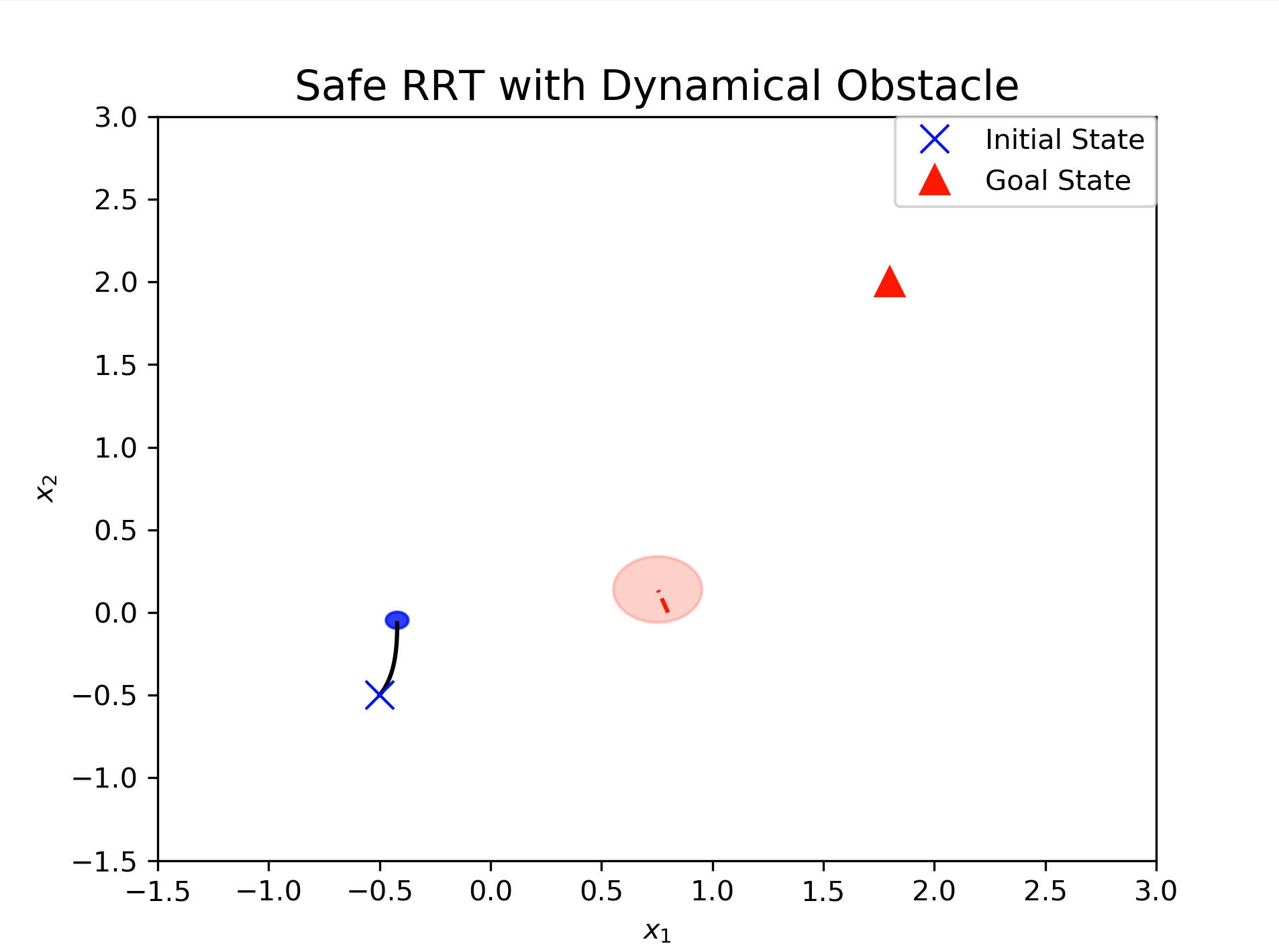}\\
\small{(a)}
\end{tabular}
\begin{tabular}{@{}c@{}}
\includegraphics[width=.40\textwidth,trim={0 0 0 2.7cm},clip=True]{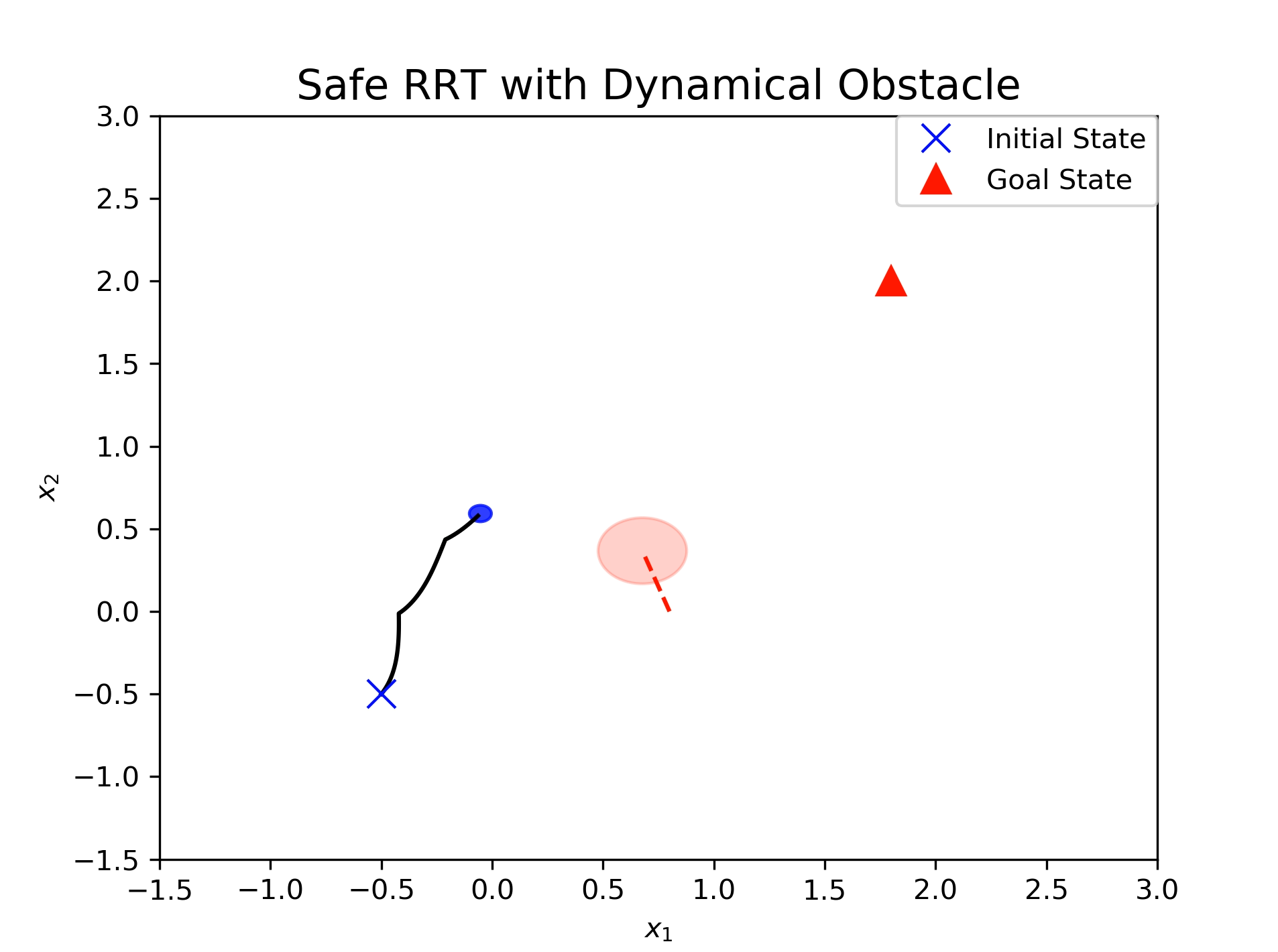}\\
\small{(b)}
\end{tabular}
\begin{tabular}{@{}c@{}}
\includegraphics[width=.40\textwidth,trim={0 0 0 2.2cm},clip=True]{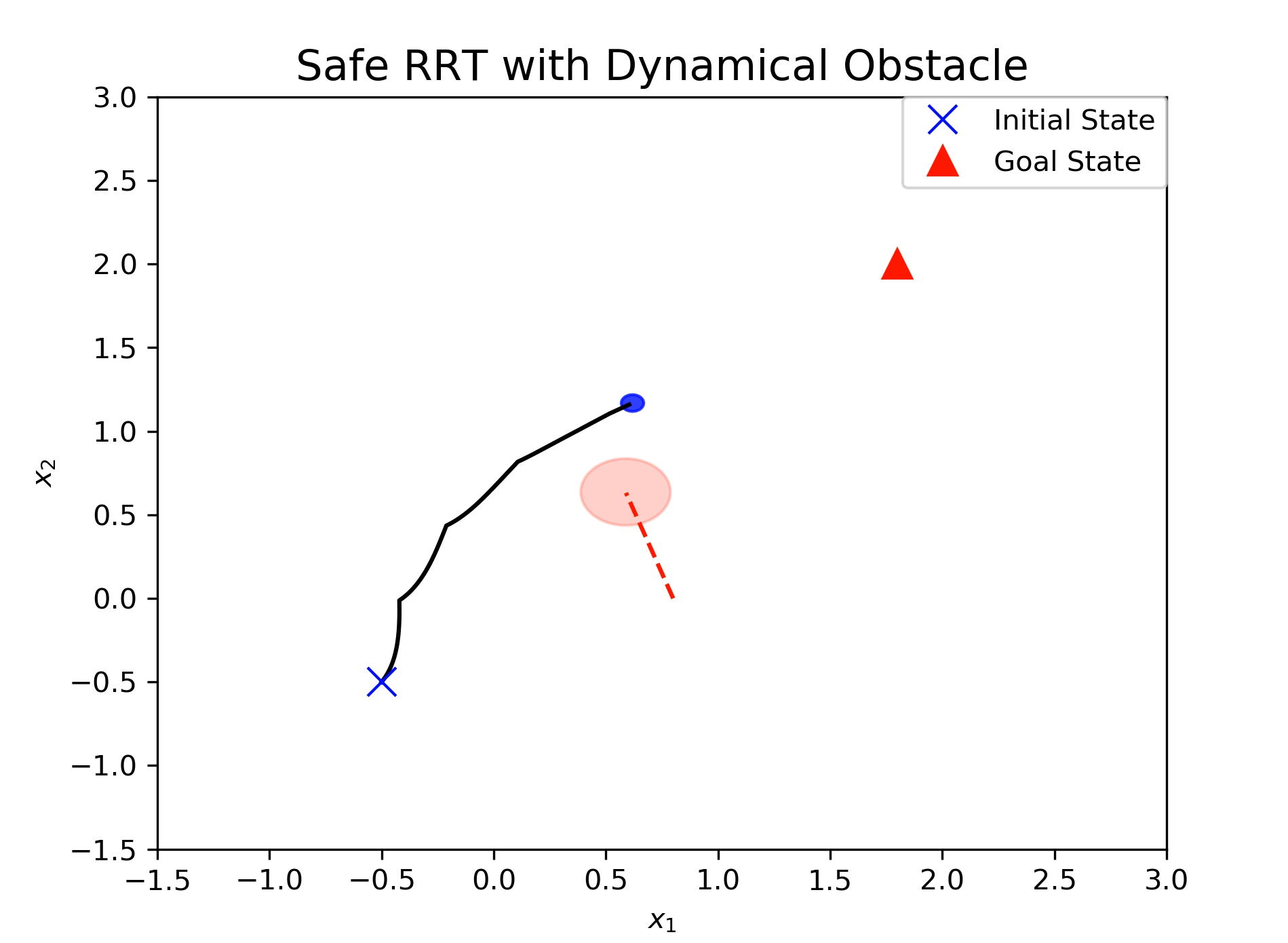}\\
\small{(c)}
\end{tabular}
\begin{tabular}{@{}c@{}}
\includegraphics[width=.40\textwidth,trim={0 0 0 2.4cm},clip=True]{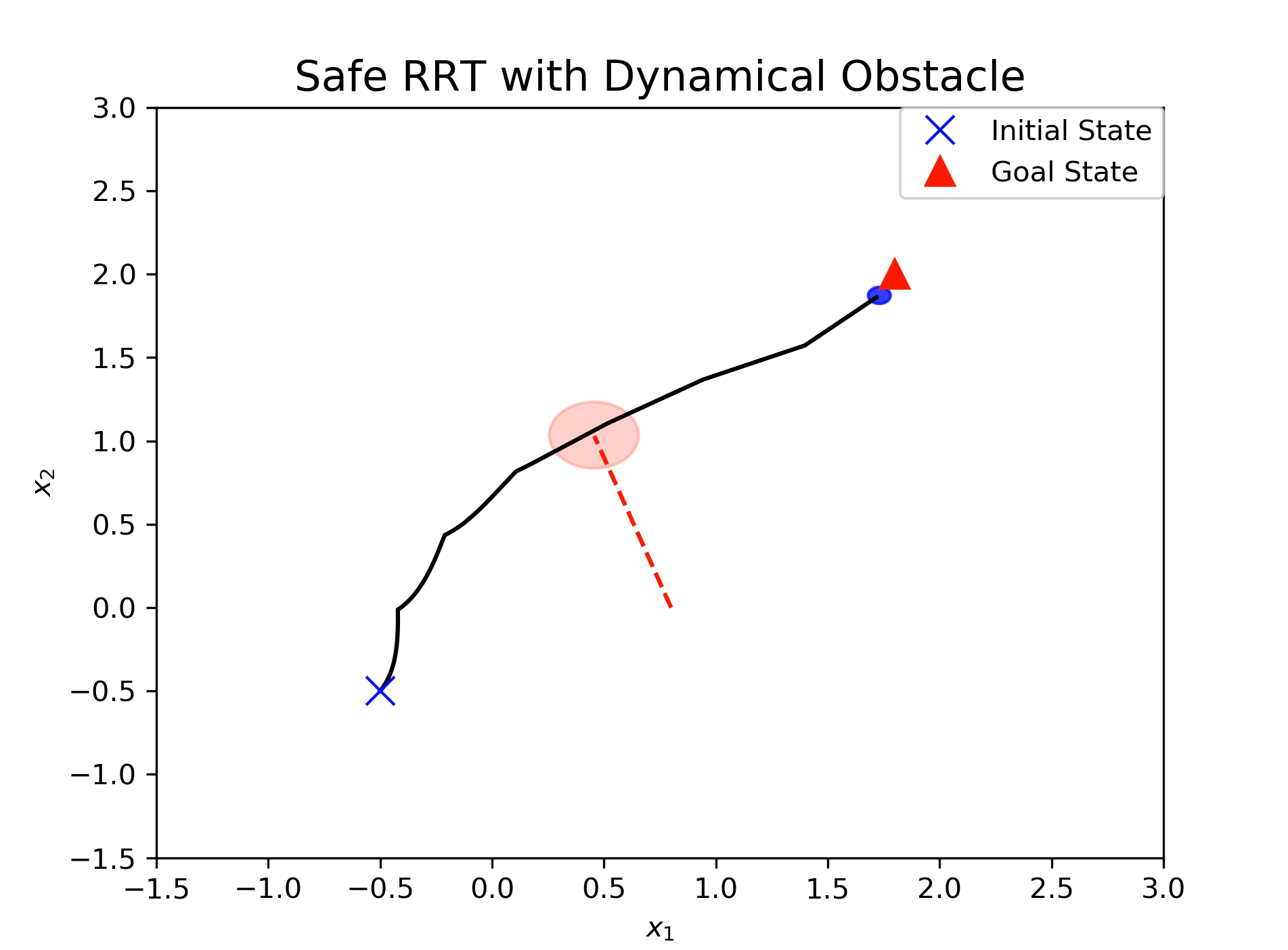}\\
\small{(d)}
\end{tabular}
\end{center}
\caption{Example 2 CBF-RRT snapshots ($\sigma^2 = 0.2$): The plots from (a) to (d) demonstrate how the robot (blue dot) progress as the obstacle (red circle) crosses its path. The \texttt{SafeSteer} function steers the robot away from the obstacle such that it can reach the goal without any collision.}\label{fig:example2snapshots}
\vspace{-5pt}
\end{figure}

In this example, without the loss of generality, we consider a single obstacle with constant velocities $v_{obs,1},v_{obs,2}$. We formulate a composite system that includes obstacle dynamics \eqref{eq:obsDynamicsExamples} as the following
\begin{align}\label{eq:compositeSys}
    \left[\begin{matrix}\dot{x_1} \\ \dot{x_2} \\ \dot{\theta}\\\dot{x}_{obs,1}\\\dot{x}_{obs,2} \end{matrix}\right]= \left[ \begin{matrix} v \cos(\theta)\\ v \sin(\theta) \\0 \\v_{obs,1}\\v_{obs,2} \end{matrix} \right]  + \left[ \begin{matrix} 0\\ 0 \\1 \\0 \\0 \end{matrix} \right] \omega.
\end{align}
\begin{remark}
The composite system \eqref{eq:compositeSys} can be easily extended to multi-agents path planning problems. In this example, we only consider path planning problem for a single agent. 
\end{remark}
\vspace{-10pt}
\begin{figure}[H]
\begin{center}
\begin{tabular}{@{}c@{}}
\includegraphics[width=.45\textwidth,trim={0 0 0 1.3cm},clip=True]{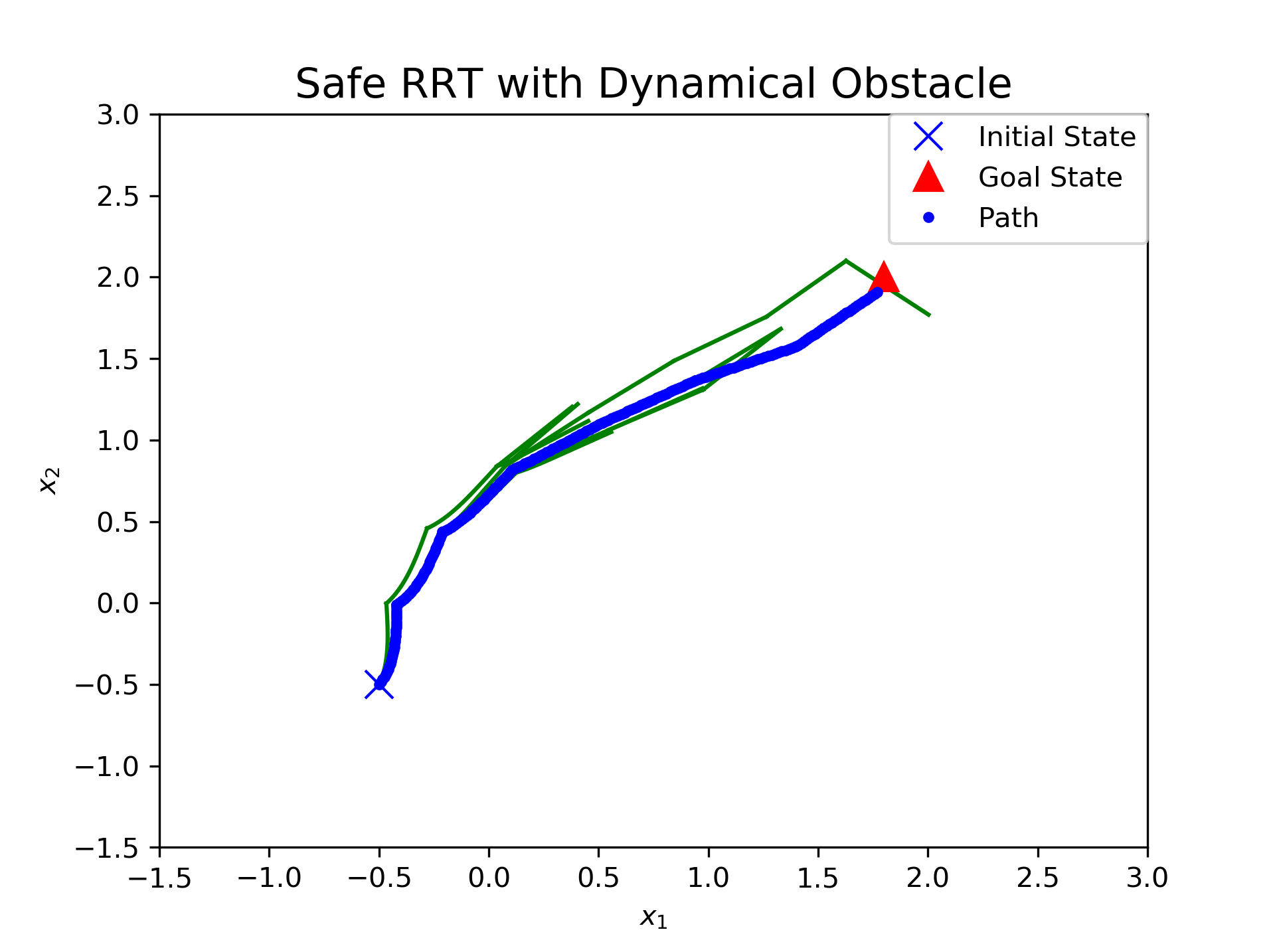}\\
\small{Variance $\sigma^2 = 0.2$}
\end{tabular}
\begin{tabular}{@{}c@{}}
\includegraphics[width=.45\textwidth,trim={0 0 0 1.3cm},clip=True]{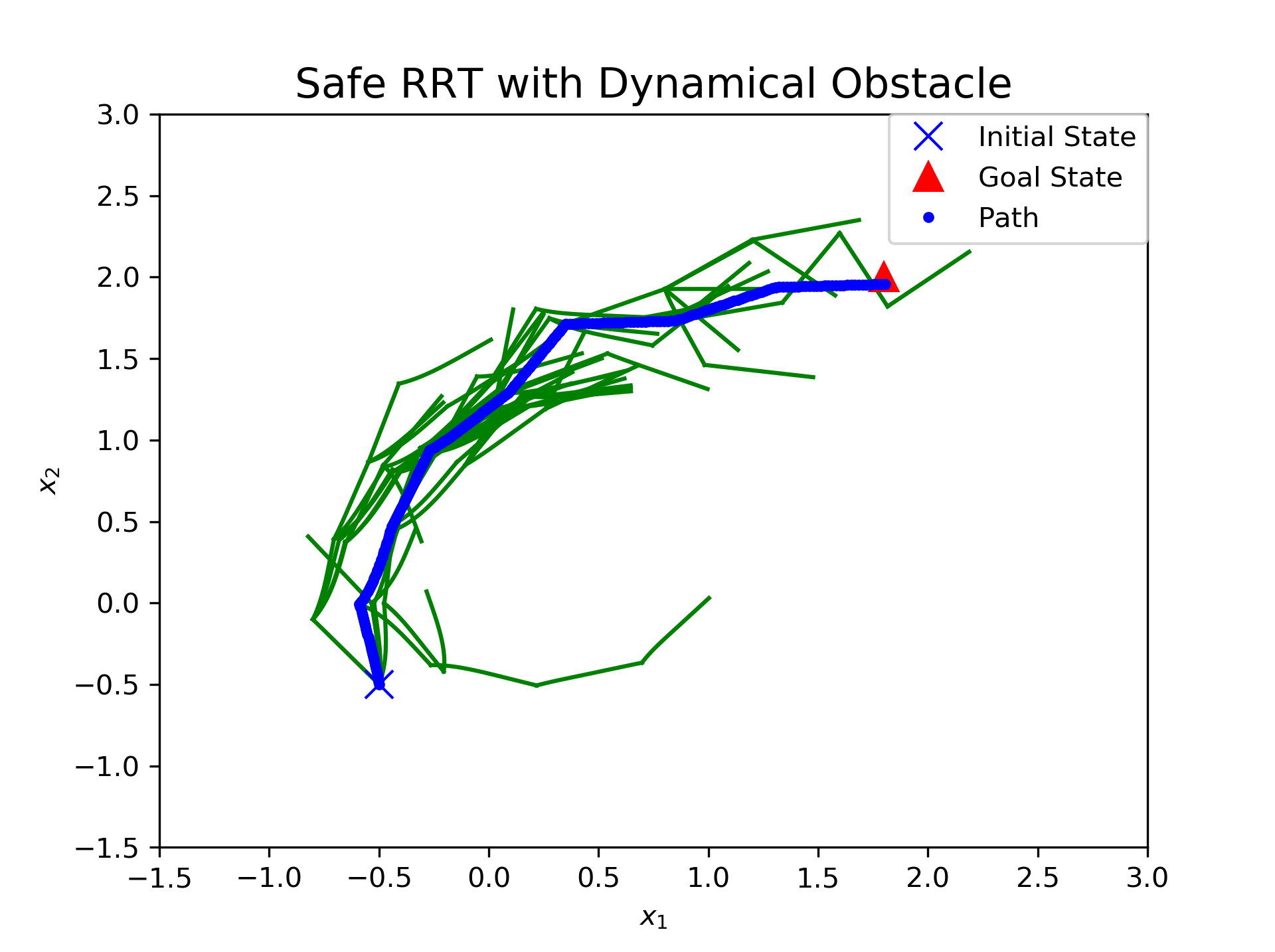}\\
\small{Variance $\sigma^2 = 0.6$}
\end{tabular}
\end{center}
\caption{Example 2: CBF-RRT generated tree structure (Obstacle is not included)}\label{fig:Example2Tree}
\vspace{-5pt}
\end{figure}
We enforce the ECBF constraint \eqref{eq:generalECBFexample} and the result from Fig. \ref{fig:Example2Tree} shows the generated path under different variances. Similar to the result from example 1, the case with $\sigma^2 = 0.6$ leads to a more sparse tree and both trajectories are able to avoid the moving obstacle as shown in Fig. \ref{fig:example2snapshots}. The hyper parameters for this example is shown in Table \ref{tb:example2}.
\begin{table}
\centering
\caption{Hyper Parameters for Example 2}\label{tb:example2}
\begin{tabular}{ |c|c|c|c|c|c|c|c|c|c| } 
\hline
 Case&$k_1$ & $k_2$ & $\sigma^2$  & $\epsilon$ & $t_h$& v &$v_{obs,1}$&$v_{obs,2}$& Run Time \\
\hline
 1&0.6 & 1.5 & 0.2& 0.15 &0.5 &1.0 &-0.1 &0.3 &17.62s \\ 
 \hline
  2&0.6 & 1.5 & 0.6& 0.15 &0.5 &1.0 &-0.1 &0.3 &2.48s\\
\hline
\end{tabular}
\end{table}
\vspace{-10pt}
\subsection{Comparison with RRT and RRT* End-point Collision Check}
\begin{figure}[H]
\begin{center}
\begin{tabular}{@{}c@{}}
\includegraphics[width=.30\textwidth,trim={0 0 0 3.3cm},clip=True]{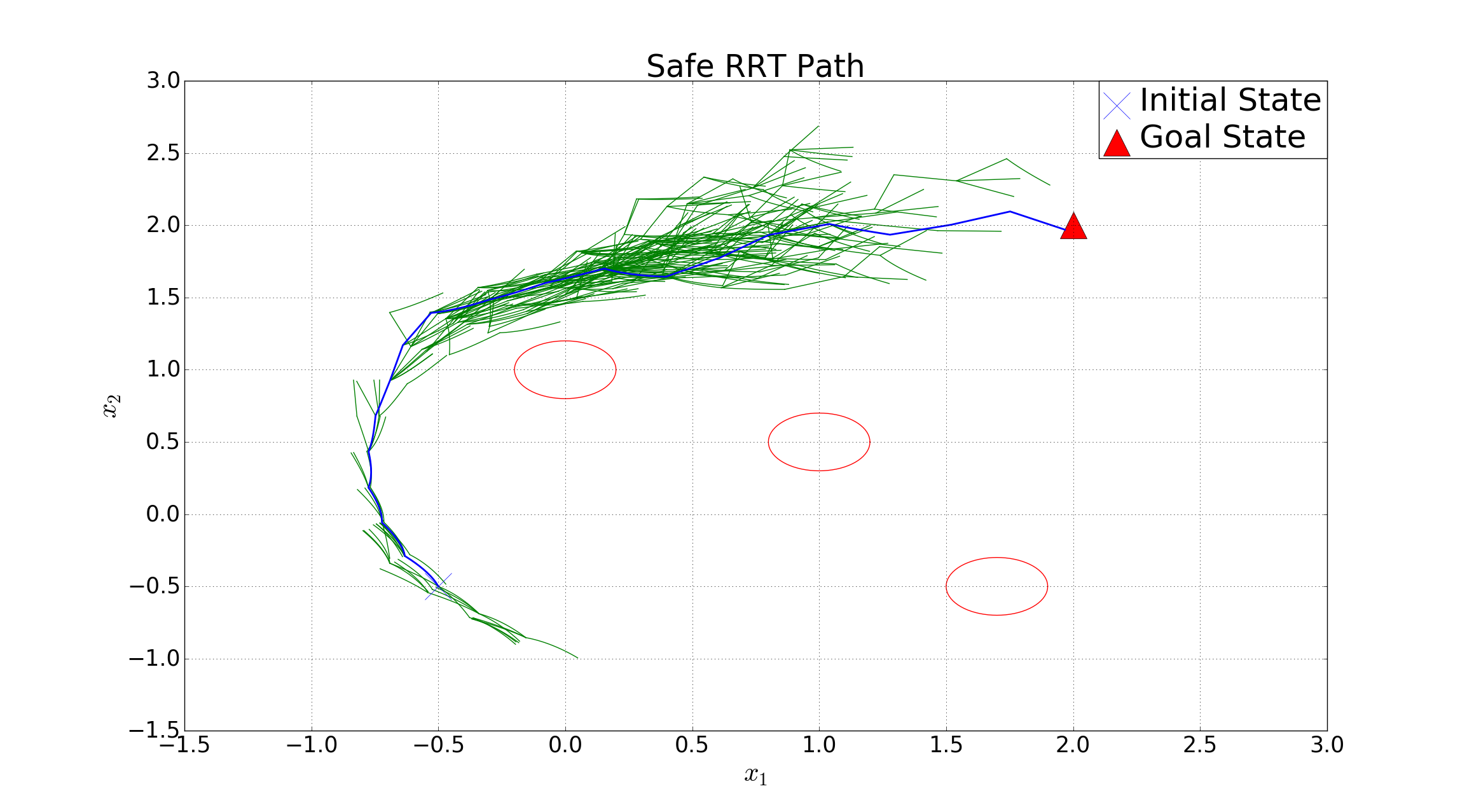}\\
\small{(a)}
\end{tabular}
\begin{tabular}{@{}c@{}}
\includegraphics[width=.30\textwidth,trim={0 0 0 3.3cm},clip=True]{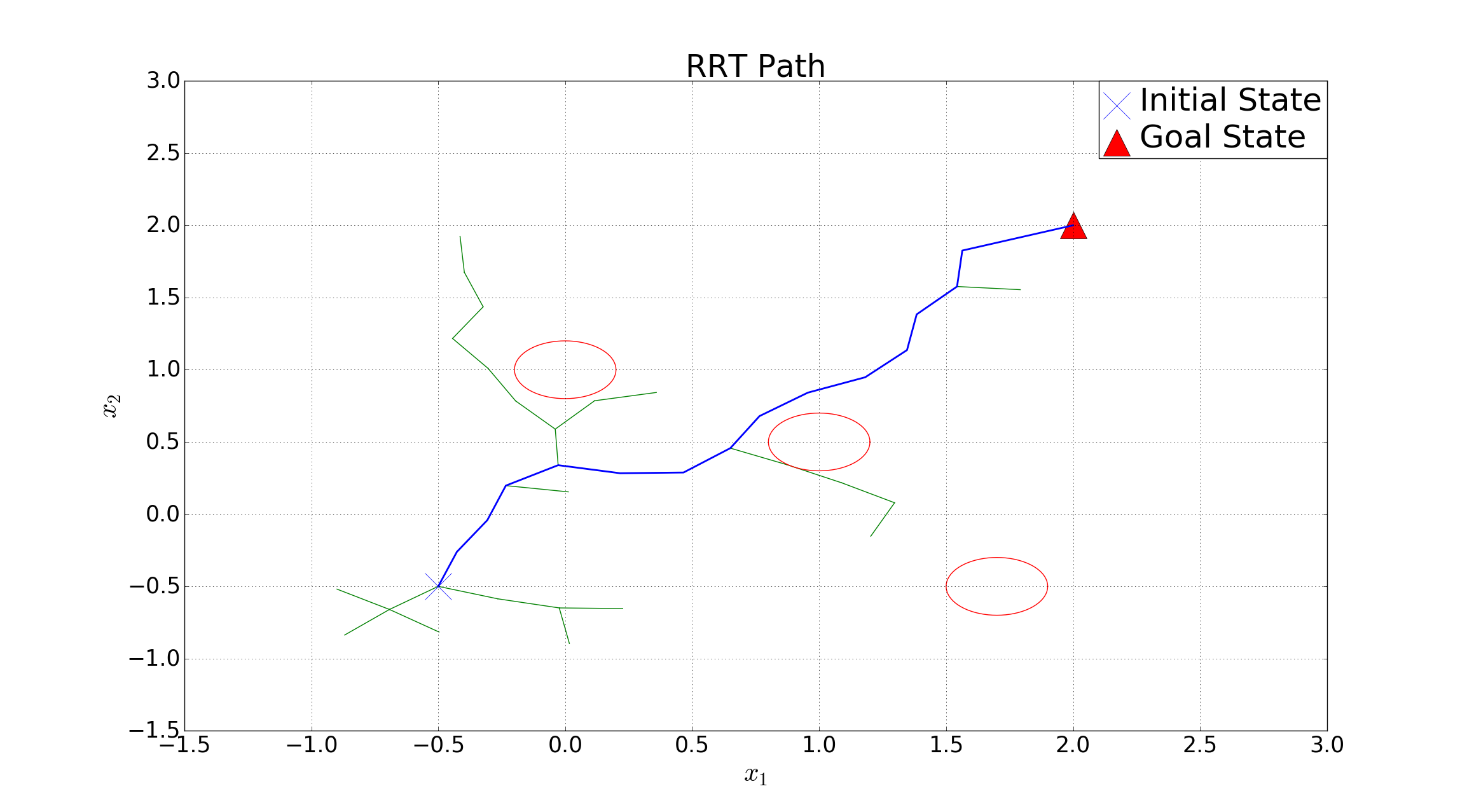}\\
\small{(b)}
\end{tabular}
\begin{tabular}{@{}c@{}}
\includegraphics[width=.30\textwidth,trim={0 0 0 3.3cm},clip=True]{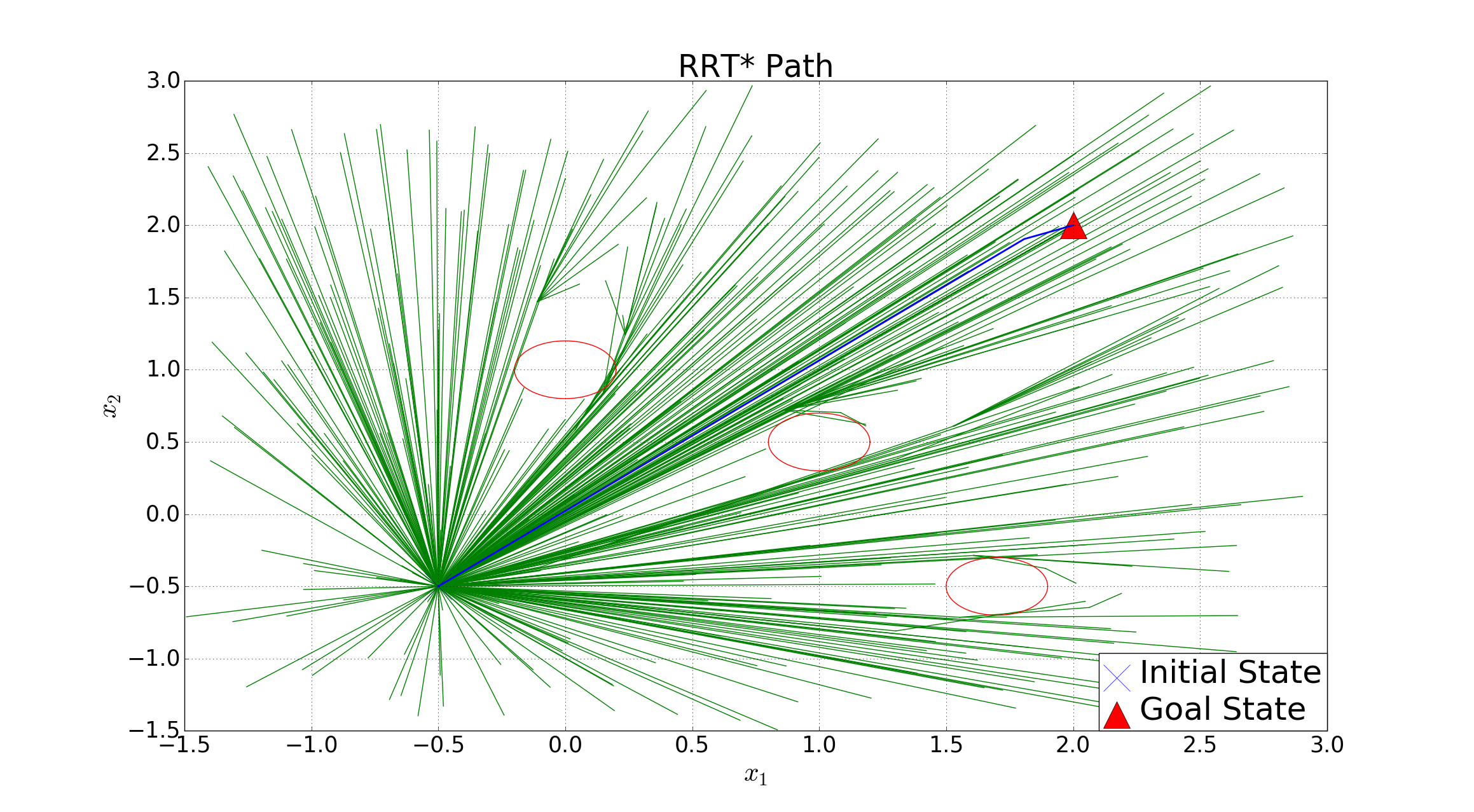}\\
\small{(c)}
\end{tabular}
\begin{tabular}{@{}c@{}}
\includegraphics[width=.30\textwidth,trim={0 0 0 3.3cm},clip=True]{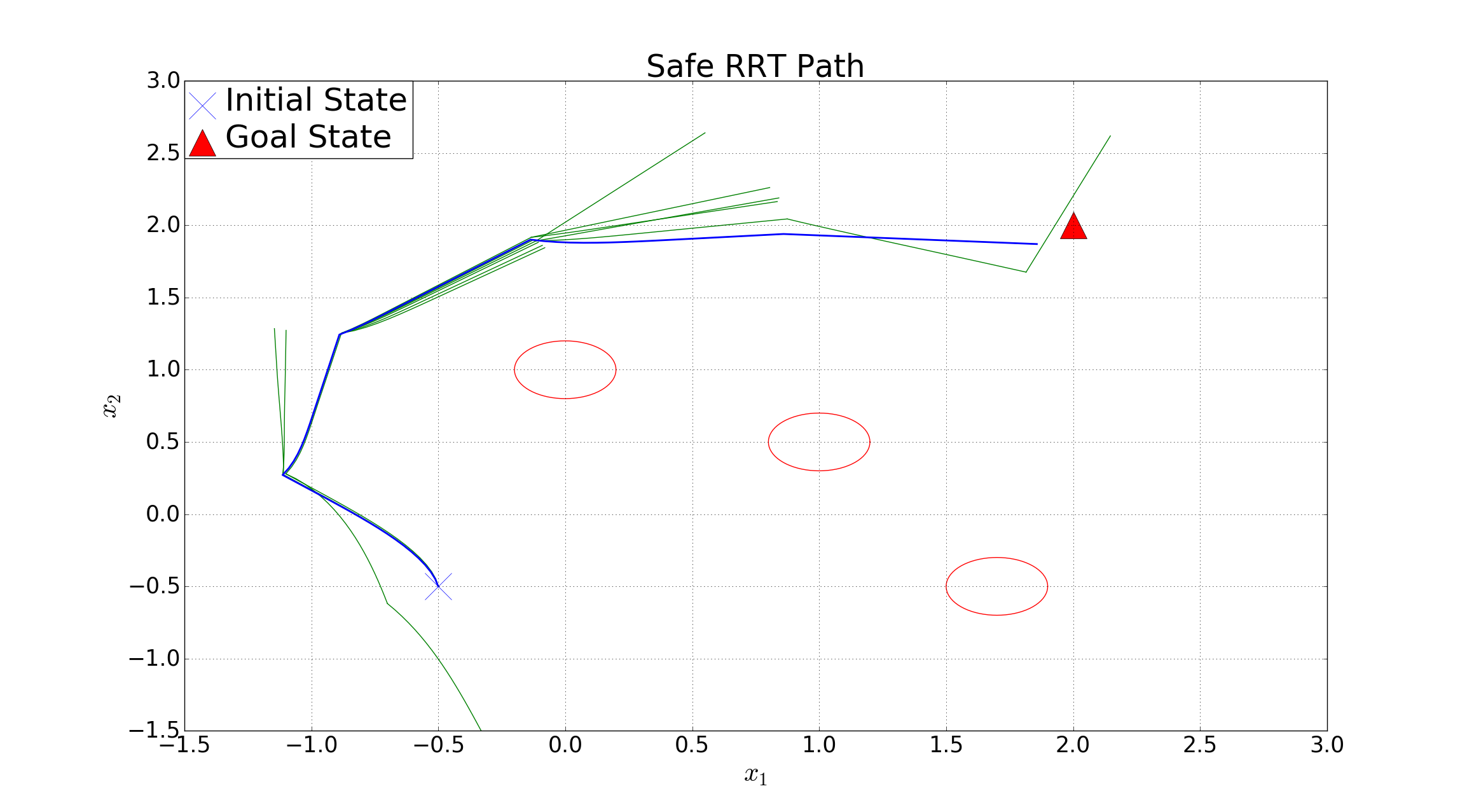}\\
\small{(d)}
\end{tabular}
\begin{tabular}{@{}c@{}}
\includegraphics[width=.30\textwidth,trim={0 0 0 3.3cm},clip=True]{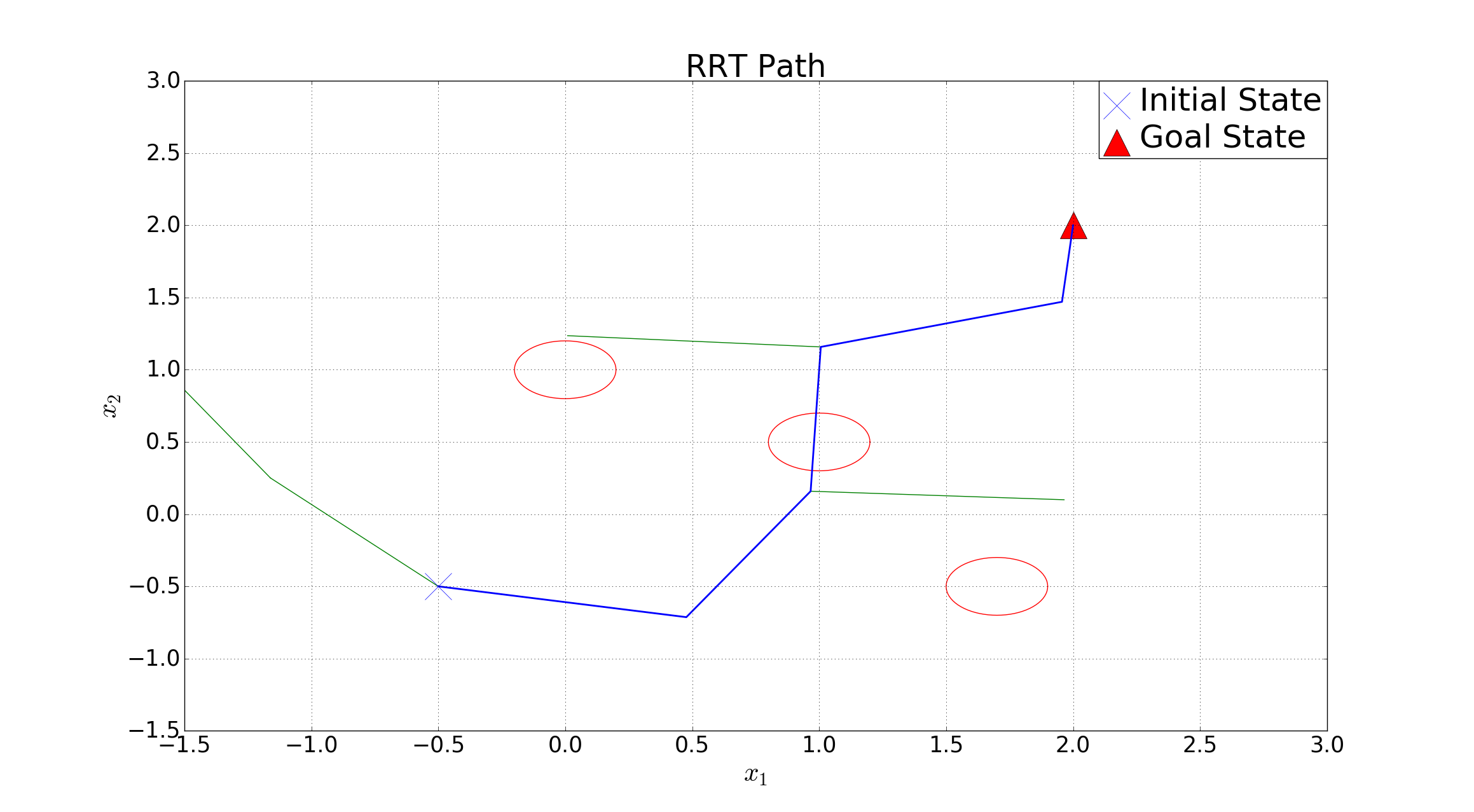}\\
\small{(e)}
\end{tabular}
\begin{tabular}{@{}c@{}}
\includegraphics[width=.30\textwidth,trim={0 0 0 3.3cm},clip=True]{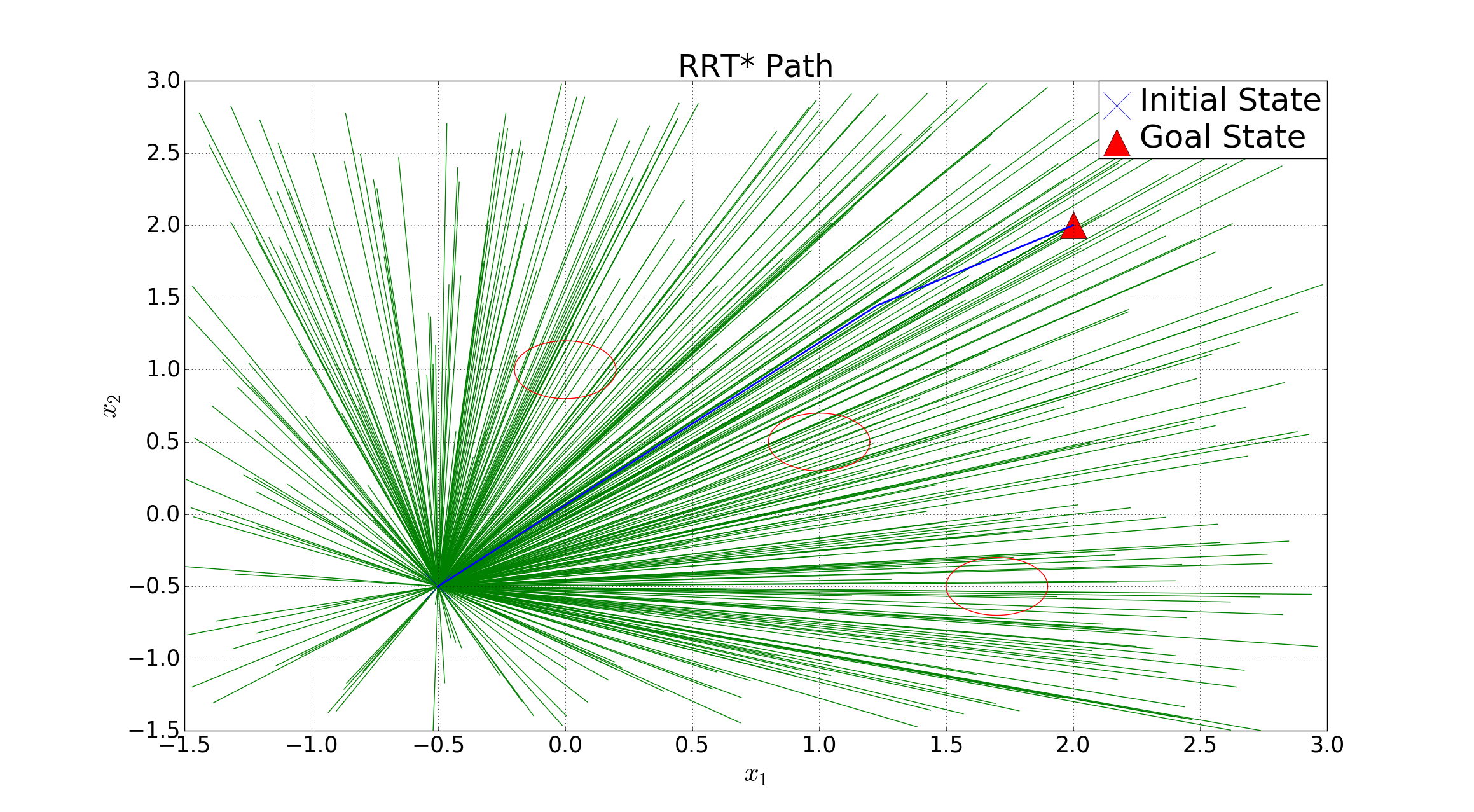}\\
\small{(f)}
\end{tabular}
\end{center}
\vspace{-5pt}
\caption{Comparison between CBF-RRT, RRT and RRT* with End-point collision check. (a) CBF-RRT with $t_h = 0.25s$. (b) RRT with $\delta d = 0.25m$. (c) RRT* with $\delta d = 0.25m$. (d) CBF-RRT with $t_h = 1.00s$. (e) RRT with $\delta d = 1.00m$. (f) RRT* with $\delta d = 1.00m$.}\label{fig:RRTcomparison}
\vspace{-5pt}
\end{figure}

One of the major advantages of CBF-RRT is the guarantee of collision free trajectory generation in continuous time. In RRT and RRT* path planning, an end-point collision check function typically only checks if the end vertex $x_{new}$ is within an obstacle region. While it is computationally efficient, the generated trajectories have the potential to collide with intermediary obstacles that are smaller than the chosen step size $\delta d$ (Fig. \ref{fig:classicRRT}). We compare our proposed CBF-RRT algorithm with both RRT and RRT* in the same static environment from Example 1 and Fig. \ref{fig:RRTcomparison} highlights how trajectories have the potential to collide with the obstacles when $\delta d$ is chosen to be too large. CBF-RRT however does not suffer from this issue since it considers continuous trajectories and has no such explicit collision check. Table \ref{tb:example3} shows the run time and number of vertices comparisons between the CBF-RRT, RRT, and RRT* algorithms. CBF-RRT preforms between RRT and RRT* in terms of both computation time and number of vertices generated. This illustrates that CBF-RRT is able to guarantee collision free trajectories for non-linear dynamics without excessive increases in computation time over less-safe RRT and RRT*. Furthermore, we expect CBF-RRT to outperform RRT and RRT* in highly complex (crowded) environments because it does not need to perform a collision check explicitly.  
\vspace{-10pt}
\vspace{-10pt}
\begin{table}
\centering
\caption{Comparison of CBF-RRT, RRT, and RRT*}\label{tb:example3}
\begin{tabular}{ |c|c|c|c| } 
\hline
 Algorithm& Step Size&Run Time (s)& Number of Vertices \\
\hline
  CBF-RRT & 0.25s & 2.98 & 496\\
  \hline
  RRT & 0.25m & 0.00125 & 39\\
  \hline
  RRT* & 0.25m & 2.109 & 488\\
  \hline
  CBF-RRT & 1s& 0.281 & 26 \\ 
 \hline
  RRT & 1m & 0.00027 & 9\\
  \hline
  RRT* & 1m & 1.480 & 494\\
  \hline
\end{tabular}
\end{table}
\vspace{-5pt}
\section{Conclusion and Future Work}\label{sec:conclusion}
In this paper we present CBF-RRT, a motion planning algorithm that successfully generates collision-free trajectories and control strategies for a nonlinear system under both static and dynamic environments. In future work, we will further increase the richness of the mission specifications by adding temporal logic based constraints. Second, we will further improve the current algorithm with rewiring technique that is similar to RRT* and perform theoretical analysis. Third, we would like to extend the current algorithm to handle multi-agent motion planning.
\vspace{-5pt}
\section{Acknowledgements}
This work was partially supported by MIT Lincoln Labs 7000443517 and NSF IIS-1723995.

\vspace{1.0cm}
\bibliographystyle{ieeetr}
\bibliography{myBib}

\end{document}